\documentclass[10pt,twocolumn,letterpaper]{article}

\usepackage{iccv}
\usepackage{times}
\usepackage{epsfig}
\usepackage{graphicx}
\usepackage{amsmath}
\usepackage{amssymb}

\usepackage{booktabs}
\usepackage{multirow}
\usepackage{tabularx}
\usepackage{caption}
\usepackage{subcaption}
\usepackage{makecell}
\usepackage{threeparttable}
\usepackage{balance}
\usepackage{floatrow}
\newfloatcommand{capbtabbox}{table}[][\FBwidth]

\usepackage[pagebackref=true,breaklinks=true,letterpaper=true,colorlinks,bookmarks=false]{hyperref}
\usepackage{xr}
\usepackage{mathtools}
\newcommand{\defeq}{\vcentcolon=}
\newcommand{\framewidth}{0.24\linewidth}
\newcommand{\figwidth}{0.24\linewidth}
\newcommand{\newfigwidth}{0.485\linewidth}
\usepackage{array}
\newcolumntype{?}{!{\vrule width 1pt}}

\iccvfinalcopy 

\def\iccvPaperID{5471} 

\begin{document}

\title{Uncertainty-aware State Space Transformer for Egocentric \\ 3D Hand Trajectory Forecasting}

\author{
Wentao Bao $^{1*}$, 
Lele Chen$^{2}$, 
Libing Zeng$^{3*}$, 
Zhong Li$^{2}$, 
Yi Xu$^{2}$, 
Junsong Yuan$^{4}$,  
Yu Kong$^{1}$ \\
$^{1}$Michigan State University, 
$^{2}$OPPO US Research Center,\\
$^{3}$Texas A\&M University,
$^{4}$University at Buffalo\\
{\tt\small \{baowenta,yukong\}@msu.edu, libingzeng@tamu.edu,} \\
{\tt\small \{lele.chen,zhong.li,yi.xu\}@oppo.com, jsyuan@buffalo.edu}
}

\twocolumn[{%
\renewcommand\twocolumn[1][]{#1}%
\maketitle
\ificcvfinal\thispagestyle{empty}\fi

\begin{center}
\vspace{-5mm}
    \centering
    \captionsetup{type=figure}
    \includegraphics[width=\textwidth]{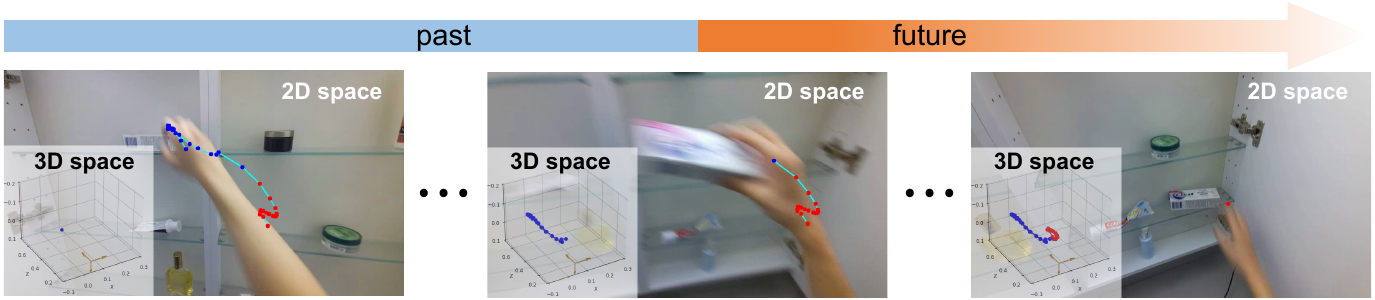}
    \captionof{figure}{\small\textbf{Egocentric 3D Hand Trajectory Forecasting}. Our goal is to predict the future 3D hand trajectory (in \textcolor{red}{\textbf{red}}) given the past observations of egocentric video and trajectory (in \textcolor{blue}{\textbf{blue}}). Compared to the 2D image space, predicting trajectory in a global 3D space is practically more valuable to understand human intention and behavior in AR/VR applications.}
    \label{fig:task}
\end{center}
}]

\def\thefootnote{*}\footnotetext{This work is the internship project of Wentao Bao and Libing Zeng mentored by Lele Chen at OPPO US Research Center.}

\begin{abstract}
   Hand trajectory forecasting from egocentric views is crucial for enabling a prompt understanding of human intentions when interacting with AR/VR systems. However, existing methods handle this problem in a 2D image space which is inadequate for 3D real-world applications. In this paper, we set up an egocentric 3D hand trajectory forecasting task that aims to predict hand trajectories in a 3D space from early observed RGB videos in a first-person view. To fulfill this goal, we propose an uncertainty-aware state space Transformer (USST) that takes the merits of the attention mechanism and aleatoric uncertainty within the framework of the classical state-space model. The model can be further enhanced by the velocity constraint and visual prompt tuning (VPT) on large vision transformers. Moreover, we develop an annotation workflow to collect 3D hand trajectories with high quality. Experimental results on H2O and EgoPAT3D datasets demonstrate the superiority of USST for both 2D and 3D trajectory forecasting. The code and datasets are publicly released: {\small{\url{https://actionlab-cv.github.io/EgoHandTrajPred}}}.
\end{abstract}

\section{Introduction}
\label{sec:intro}

Egocentric video understanding aims to understand the camera wearers' behavior from the first-person view. It is receiving increasing attention in recent years~\cite{Ego4D_CVPR2022,EgoPAT3D_CVPR2022,Girdhar_ICCV2021,liu2021egocentric,li2021ego,nagarajan2020ego,wang2020symbiotic,liu2020forecasting,EK100_PAMI2020,EK50_ECCV2018,Li_ECCV2018} due to its analogousness to the way human visually perceives the world. An important egocentric vision task is to forecast the egocentric hand trajectory of the camera wearer~\cite{OCT_CVPR2022}, which has great value in Augmented/Virtual Reality (AR/VR) applications. For example, the predicted 3D trajectories can help plan and stabilize a patient’s 3D hand motion who has upper-limb neuromuscular disease~\cite{EgoPAT3D_CVPR2022}. Besides, the early predicted 3D hand trajectory is key to reducing rendering latency in VR games for achieving an immersive gaming experience~\cite{gamage2021so}.

In the existing literature, egocentric 3D hand trajectory forecasting is far from being explored. The method in~\cite{OCT_CVPR2022} could only predict 2D trajectory on an image and cannot forecast precise 3D hand movements. Recent works~\cite{Bi_ECCV2020,wang2021multi,yuan2019ego} predict the trajectory or 3D human motions from egocentric views, but they do not predict the 3D trajectory of the camera wearer. Besides, though forecasting the 3D hand pose provides fine-grained information about 3D hands~\cite{diller2022forecasting}, it is out of our scope as we focus on the camera wearers' planning behavior revealed by 3D hand trajectory.

The challenges of egocentric video-based 3D trajectory forecasting are significant. First, accurate large-scale 3D trajectory annotations are labor-intensive and expensive. They rely on wearable markers or multi-camera systems for hand motion capture in a controlled environment.
Second, learning the depth of 3D trajectory from egocentric videos is challenging. On one hand, using 2D video frames to estimate 3D trajectory depth is an ill-conditioned problem similar to other monocular 3D tasks~\cite{Tekin_2019_CVPR,panteleris2018using,bao2019monofenet}. Even if the historical 3D hand trajectory is utilized as the input, how to exploit the visual and trajectory information for forecasting is still nontrivial. On the other hand, due to the inevitable camera motion in an egocentric view, the background of the scene is visually dynamic which poses a significant barrier to inferring the foreground depth~\cite{li2019learning,zhang2021consistent}. Third, as a Seq2Seq forecasting problem, it is critical to formulate the latent transition dynamics~\cite{girin2021dynamical,bao2020uncertainty} that allows the variances of data due to anytime forecasting and limited observations.

In this paper, we address these challenges by developing an uncertainty-aware state space transformer (USST). It follows the state-space model~\cite{rangapuram2018deep} by taking the observed egocentric RGB videos with the historical 3D trajectory as input to predict future 3D trajectory. 
Our model deals with the depth noise of trajectory annotation by introducing the depth robust aleatoric uncertainty in training. To fuse the information from the dense RGB pixels and sparse historical trajectory, we leverage visual and temporal transformer encoders as backbones and utilize the recent visual prompt tuning (VPT) to enhance the visual features. Following the state space model, we develop a novel attention-based state transition module and an emission module with a predictive link to predict the 3D global trajectory coordinates. Moreover, to take the hand motion inertia into consideration, we propose a velocity constraint to regularize the model training, which helps generalize to unseen scenarios.

To enable egocentric 3D hand trajectory prediction, we follow~\cite{EgoPAT3D_CVPR2022} to develop a scalable annotation workflow to automate the annotation on RGB-D data from head-mounted Kinect devices. 
In particular, camera motion is estimated to transform the 3D trajectory annotations from local to global camera coordinate system. Experimental results on H2O and EgoPAT3D datasets show that our method is effective and superior to existing Transformer-based approaches~\cite{OCT_CVPR2022} and other general Seq2Seq models. In summary, our contributions are as follows:
\begin{itemize}
    \item We propose an uncertainty-aware state space transformer (USST) that consists of a novel state transition and emission, aleatoric uncertainty, and visual prompt tuning, which are empirically found effective.
    \item We collected and will release our annotations on H2O~\cite{H2O_ICCV21} and EgoPAT3D~\cite{EgoPAT3D_CVPR2022} datasets that will benefit the egocentric 3D hand trajectory forecasting research.
    \item We benchmarked recent methods on the proposed task and experimental results show that our method achieves the best performance and could be generalizable to unseen egocentric scenes.
\end{itemize}

\section{Related Work}
\label{sec:related_work}

\paragraph{Trajectory Prediction} Predicting the physical trajectory of moving objects is a long-standing research topic. It has been widely studied in applications for pedestrians~\cite{alahi2016social,zhang2019sr}, vehicle~\cite{deo2018convolutional,malla2022nemo}. Many of them are developed for the third-person view and predict trajectory in 2D pixel space. Given that the first-person view is more realistic in AR/VR applications, recent few works~\cite{OCT_CVPR2022,FHOI_ECCV2020} are trying to predict the hand-object interaction from egocentric videos. Though the method in~\cite{Bi_ECCV2020} predicts the pedestrian trajectory in 3D space, their method primarily addresses the social interaction of multiple pedestrians. The recent work~\cite{qiu2022egocentric} also targets the egocentric 3D trajectory for pedestrian scenarios, but their method leverages depth modality and nearby person's trajectory as context information which are practically uneasy to collect. Besides, due to the annotation noise and uncertain nature of trajectory prediction, probabilistic modeling is widely adopted in existing literature~\cite{wiest2012probabilistic,thiede2019analyzing,malla2022nemo}. In this paper, following the probabilistic setting, we step toward the egocentric 3D hand trajectory prediction using practically accessible RGB videos for AR/VR scenarios.

\begin{figure*}[t]
    \centering
    \includegraphics[width=\textwidth]{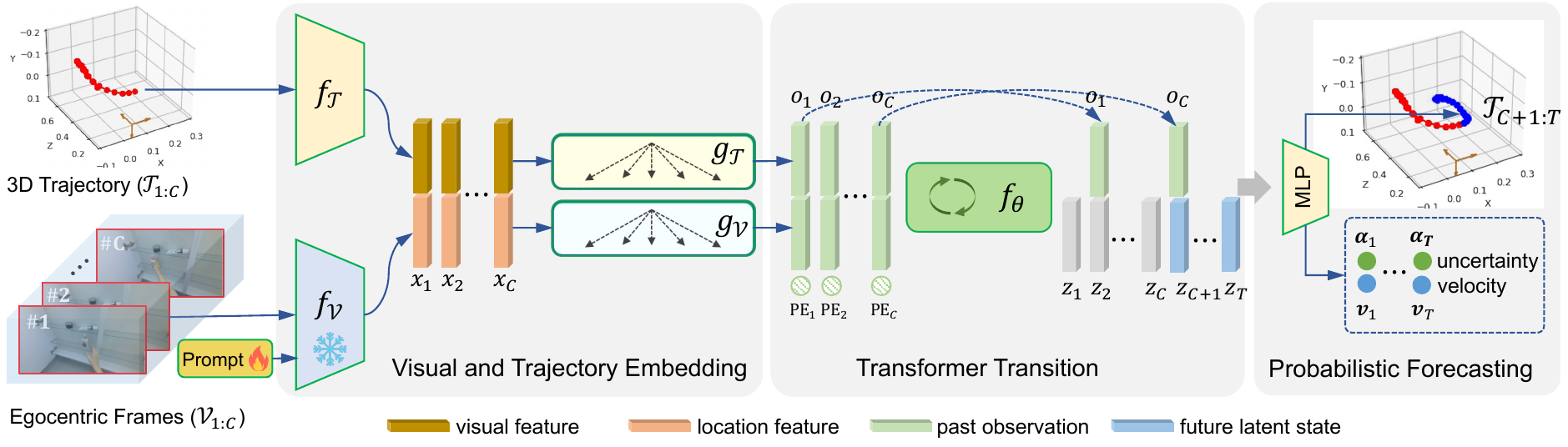}
    \caption{\small\textbf{Proposed USST Model}. Given the RGB frames and 3D hand locations of $C$ observed time steps, we extract and concatenate their features as $\mathbf{x}_{1:C}$ by the prompted backbone $f_\mathcal{V}$ and MLP model $f_{\mathcal{T}}$, which are further fed into transformer encoders to produce temporal observations $\mathbf{o}_{1:C}$. Together with positional encodings $\texttt{PE}_{1:T}$ of the full horizon, our state transition layer could recursively extrapolate the latent states $\mathbf{z}_{C+1:T}$ for 3D trajectory forecasting along with uncertainty and velocity ($\boldsymbol{\alpha}$ and $\mathbf{v}$) in $T-C$ future steps.}
    \label{fig:framework}
\end{figure*}

\vspace{-2mm}
\paragraph{Egocentric Video Representation} Egocentric videos are recorded in a first-person view. Different from the videos in a third-person view, learning an egocentric video representation is more challenging due to the dynamic background caused by camera motion and implicit intention of activities from camera wearers~\cite{li2021ego,fathi2011understanding,nunez2022egocentric}.
For 3D trajectory prediction, existing commonly used egocentric video datasets such as EPIC-Kitchens~\cite{EK50_ECCV2018,EK100_PAMI2020} do not provide the depth information and camera parameters, which are essential for annotating the 3D hand trajectory. Though the recent Ego4D~\cite{Ego4D_CVPR2022} benchmark provides a hand forecasting subset, the annotations are defined as 2D locations in image space. Therefore, we resort to a cost-effective workflow to collect 3D hand trajectory annotations from existing 3D hand pose datasets such as the EgoPAT3D~\cite{EgoPAT3D_CVPR2022} and H2O~\cite{H2O_ICCV21} datasets. Our annotation workflow can be deployed to any egocentric dataset collected by head-mounted RGB-D sensors.

\vspace{-2mm}
\paragraph{State Space Model} State-space Model (SSM) originates from the control engineering field. It is conceptually general and inspires many classical SSMs such as the Kalman Filtering for prediction tasks. Recent deep SSMs~\cite{STORN_NIPSW2014,VRNN_NIPS2015,SRNN_NIPS2016} are increasingly popular by combining Recurrent Neural Networks (RNNs) with the Variational AutoEcoders (VAE). However, these approaches are limited in practice due to the complex long-term dependency on highly-structured sequential data. In~\cite{rangapuram2018deep}, a deep SSM is proposed by combining Kalman Filtering with deep neural networks. However, the linear Gaussian assumption is uneasy to hold for high-dimensional data in the real world. To address these challenges, ProTran~\cite{ProTran_NIPS2021} introduces a probabilistic Transformer~\cite{Transformer_NIPS2017} under a variational SSM for time-series prediction. However, maximizing the variational low bound of ProTran suffers from notorious KL vanishing issue~\cite{Fu_NACCL2019}. AgentFormer~\cite{AGF_ICCV21} can also be regarded as a Transformer-based SSM, but its autoregressive decoding limits its efficiency to low-dimensional motion data. In this paper, we propose a Transformer-based SSM that allows for long-term dependency and latent dynamics efficiently in practice.

\section{Proposed Method}
\label{sec:method}

\paragraph{Problem Setup}
As shown in Fig.~\ref{fig:task}, the Egocentric 3D hand trajectory forecasting model takes as input $C$ observed RGB frames $\mathcal{V}_{1:C}=\{\mathbf{I}_1,\ldots,\mathbf{I}_C\}$ and 3D hand trajectory $\mathcal{T}_{1:C}=\{\mathbf{p}_1,\ldots,\mathbf{p}_C\}$ to predict the future 3D hand trajectory $\mathcal{T}_{C+1:T}=\{\mathbf{p}_{C+1},\ldots,\mathbf{p}_T\}$ in a finite horizon $T$. Here, $\mathbf{I}_t\in \mathbb{R}^{H\times W\times 3}$ and $\mathbf{p}_t=[x_t, y_t, z_t]^\top$ are egocentric RGB frame and 3D hand trajectory point at time step $t$, respectively. In practice, the 3D point $\mathbf{p}_t$ is defined in a global 3D world coordinate system. The ultimate goal is to learn a model $\boldsymbol{\Phi}$ by maximizing the expectation of the likelihood over the training dataset $\mathcal{D}$:
\begin{equation}
    \max_{\boldsymbol{\Phi}} \mathbb{E}_{\mathcal{V},\mathcal{T}\sim\mathcal{D}} \left[p_{\boldsymbol{\Phi}}(\mathcal{T}_{C+1:T}|\mathcal{T}_{1:C}, \mathcal{V}_{1:C})\right].
\label{eq:obj}
\end{equation}

In this paper, we formulate the problem as a state-space model. In the following sections, we will introduce the proposed model in detail.

\subsection{Uncertainty-aware State Space Transformer}
Existing SSMs could formulate the probabilistic nature of trajectory prediction. However, they do not explicitly handle the data noise issue which is commonly encountered when using RGB-D sensors to get 3D trajectory annotations. To mitigate the uncertainty from data labeling, we follow the line of research~\cite{Unct_NIPS2017,he2019bounding} and propose an uncertainty-aware state space transformer (USST) to handle the dynamics of 3D hand trajectory in egocentric scenes. 

Formally, following the latent variable modeling, the probability in Eq.~\eqref{eq:obj} over a full sequence $\mathcal{T}_{1:T}$ can be factorized by introducing $T$ latent variables $\mathcal{Z}_{1:T}$:
\begin{equation}
\resizebox{0.9\linewidth}{!}{
    $p(\mathcal{T}_{1:T}|\mathcal{T}_{1:C},\mathcal{V}_{1:C}) = \int p(\mathcal{T}_{1:T}|\mathcal{Z}_{1:T}) p(\mathcal{Z}_{1:T}|\mathcal{T}_{1:C}, \mathcal{V}_{1:C}) d\mathcal{Z}$,
}
\label{eq:likelihood}
\end{equation}
where the \emph{state transition}  $p(\mathcal{Z}_{1:T}|\mathcal{T}_{1:C},\mathcal{V}_{1:C})$ and the \emph{emission} $p(\mathcal{T}_{1:T}|\mathcal{Z}_{1:T})$ are learned from data $\mathcal{D}$. Following the SSM formulation, we propose to factorize the two terms by the independency assumptions: 
\begin{equation}
\begin{split}
    & p_{\theta}(\mathcal{Z}_{1:T}|\mathcal{T}_{1:C},\mathcal{V}_{1:C}) = \prod_{t=1}^T p_{\theta}(\mathbf{z}_t|\mathbf{z}_{1:t-1}, \mathbf{p}_{1:C},\mathbf{I}_{1:C}), \\
    & p_{\phi}(\mathcal{T}_{1:T}|\mathcal{Z}_{1:T}) = \prod_{t=1}^T p_{\phi}(\mathbf{p}_t|\mathbf{z}_t,\mathbf{p}_{t-1}),
\label{eq:ssm}
\end{split}
\end{equation}
where the latent variable $\mathbf{z}_t \in \mathcal{Z}_{1:T}$ is generated by taking as input $\mathbf{z}_{t-1}$ and the previous trajectory point $\mathbf{p}_t$ and RGB frame $\mathbf{I}_t$. To address the label noise issue, we formulate the emission model $p_{\phi}(\mathbf{p}_t|\mathbf{z}_t,\mathbf{p}_{t-1})$ as a probabilistic module to learn the aleatoric uncertainty. In the following paragraphs, we will elaborate on feature embedding, state transition, and probabilistic prediction. 

\vspace{-2mm}
\paragraph{Visual and Trajectory Embedding} As advocated by~\cite{ProTran_NIPS2021,AGF_ICCV21}, Transformers are effective to capture the long-term dependency for sequential data.  Therefore, we propose to leverage visual and temporal Transformers~\cite{Transformer_NIPS2017} as encoders to learn the features from the dense RGB frames and sparse trajectory points. Specifically, we first embed the observed sequence of egocentric RGB frames and 3D trajectory by models $f_{\mathcal{V}}$ and $f_{\mathcal{T}}$, followed by modality-specific transformers $g_{\mathcal{V}}$ and $g_{\mathcal{T}}$. This process can be expressed as
\begin{equation}
\begin{split}
    & [\mathbf{x}_t^{(\mathcal{V})}, \mathbf{x}_t^{(\mathcal{T})}] = [f_{\mathcal{V}}(\mathbf{I}_t), f_{\mathcal{T}}(\mathbf{p}_t)], \\
    & \mathbf{o}_1^{(\mathcal{V})},\ldots,\mathbf{o}_C^{(\mathcal{V})} = g_\mathcal{V}(\mathbf{x}_1^{(\mathcal{V})},\ldots,\mathbf{x}_C^{(\mathcal{V})}),\\
    & \mathbf{o}_1^{(\mathcal{T})},\ldots,\mathbf{o}_C^{(\mathcal{T})} = g_\mathcal{T}(\mathbf{x}_1^{(\mathcal{T})},\ldots,\mathbf{x}_C^{(\mathcal{T})}),
\end{split}
\end{equation}
where $f_{\mathcal{V}}$ is a vision backbone, e.g., ResNet~\cite{ResNet_CVPR2016} and ViT~\cite{ViT_ICLR2020}. $f_{\mathcal{T}}$ is implemented as MLP following~\cite{OCT_CVPR2022,EgoPAT3D_CVPR2022}. $g_{\mathcal{V}}$ and $g_{\mathcal{T}}$ are transformer encoders that consist of $B$ stacked multi-head attention blocks. For each block $b$, a single-head attention block can be expressed as
\begin{equation}
    \text{Attn}(\mathbf{Q}_b,\mathbf{K}_b,\mathbf{V}_b) = \text{softmax}\left(\frac{\mathbf{Q}_b\mathbf{K}_b^\top}{\sqrt{d}}+\mathbf{M}\right)\mathbf{V}_b,
\label{eq:attn}
\end{equation}
where $\mathbf{Q}_b,\mathbf{K}_b,\mathbf{V}_b\in\mathbb{R}^{T\times d}$ are the projected query, key, and value matrices from the output of the previous block $b-1$, i.e, $[\mathbf{Q}_b;\mathbf{K}_b;\mathbf{V}_b]=[\mathbf{W}_b^{Q}\mathbf{Q}_{b-1};\mathbf{W}_b^{K}\mathbf{K}_{b-1};\mathbf{W}_b^{V}\mathbf{V}_{b-1}]$. All the $\mathbf{W}_b$ are learnable parameters. The binary mask $\mathbf{M}\in\mathbb{R}^{T\times T}$ zeros out the last $T-C$ columns and rows for the trajectory prediction problem. To capture global temporal interaction, the input of the first block $\mathbf{Q}_0$, $\mathbf{K}_0$, and $\mathbf{V}_0$ are the same as $\mathbf{x}_t + \text{PE}(t)$ where $\text{PE}(t)$ is the positional encoding for $t\in[1,T]$. 

\begin{figure}[t]
    \centering
    \includegraphics[width=0.95\linewidth]{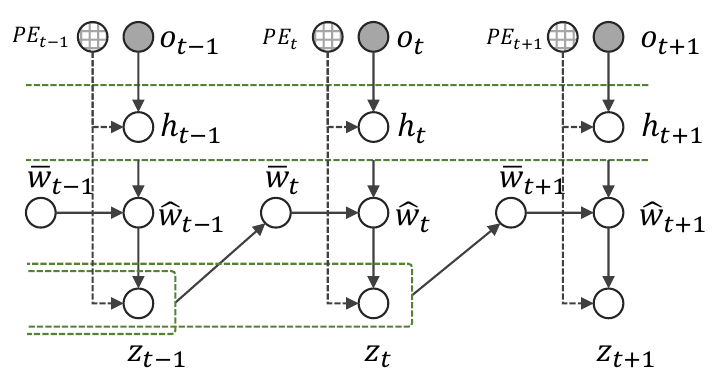}
    \captionsetup{font=small,aboveskip=5pt,belowskip=-10pt}
    \caption{\textbf{Unrolled illustration of Eq.~\eqref{eq:transition}.} Bold arrows are learnable, and green dashed lines show the attention ranges.}
    \label{supfig:transition}
\end{figure}

\vspace{-2mm}
\paragraph{Transformer Transition} With the encoded observations $\mathbf{o}_t=[\mathbf{o}_t^{(\mathcal{V})};\mathbf{o}_t^{(\mathcal{T})}]$ where we use $[;]$ to represent the feature concatenation, it is critical to formulate the state transition and future trajectory prediction based on Eq.~\eqref{eq:ssm}. Inspired by~\cite{ProTran_NIPS2021}, we propose an attention-based autoregressive module to formulate posterior $p_{\theta}(\mathbf{z}_t|\mathbf{z}_{1:t-1},\mathbf{o}_{1:C})$. Specifically, we first embed $\mathbf{o}_t$ with positional encoding by $\mathbf{h}_t = \text{LayerNorm}(\text{MLP}(\mathbf{o}_t) + \text{PE}(t))$. Then, the latent feature $\mathbf{z}_t$ is recursively encapsulated by the hidden variables $\bar{\mathbf{w}}_t$ and $\hat{\mathbf{w}}_t$ by attention modules (illustrated in Fig.~\ref{supfig:transition}):
\begin{equation}
\begin{split}
    & \bar{\mathbf{w}}_t = \text{LayerNorm}([\mathbf{z}_{t-1};\text{Attn}(\mathbf{z}_{t-1}, \mathbf{z}_{1:t-1}, \mathbf{z}_{1:t-1})]), \\
    & \hat{\mathbf{w}}_t = \text{LayerNorm}([\bar{\mathbf{w}}_t; \text{Attn}(\bar{\mathbf{w}}_t, \mathbf{h}_{1:C},\mathbf{h}_{1:C})]), \\
    & \mathbf{z}_t = \text{LayerNorm}(\text{MLP}([\hat{\mathbf{w}}_t;\text{MLP}(\hat{\mathbf{w}}_t)]) + \text{PE}(t)),
\label{eq:transition}
\end{split}
\end{equation}
where the two multi-head attention modules capture the previously generated state $\mathbf{z}_{1:t-1}$ and the hidden states of all observations $\mathbf{h}_{1:C}$. Contrary to ProTran, we use concatenation $[;]$ rather than addition before layer normalization. The insight behind this is that the queried feature from a historical context can be better preserved without being dominated by $\mathbf{z}_{t-1}$ in addition operation. Moreover, we remove the stochasticity of $\mathbf{z}_t$ and instead use a probabilistic decoder as introduced next to handle the dynamics of trajectory prediction. The benefit is that we avoid the KL divergence vanishing issue from optimizing the ELBO objective which is known to exist in variational recurrent models~\cite{Fu_NACCL2019}.

\vspace{-2mm}
\paragraph{Probabilistic Forecasting} Instead of placing the stocasticity in the transition model $p_{\theta}(\mathbf{z}_t|\mathbf{z}_{1:t-1},\mathbf{o}_{1:C})$, we propose to formulate the emission process $p_{\phi}(\mathbf{p}_t|\mathbf{z}_t, \mathbf{p}_{t-1})$ as a probabilistic model by predicting both the mean $\hat{\mathbf{p}}_t$ and variance $\hat{\boldsymbol{\sigma}}^2_t$ of each 3D hand trajectory point:
\begin{equation}
    [\hat{\mathbf{p}_t}, \hat{\boldsymbol{\alpha}}_t] = [\text{MLP}([\mathbf{z}_t;\mathbf{o}_{t-1}^{(\mathcal{T})}]), \text{softplus}(\text{MLP}([\mathbf{z}_t;\mathbf{o}_{t-1}^{(\mathcal{T})}]))],
\label{eq:emission}
\end{equation}
where the uncertainty $\hat{\boldsymbol{\alpha}}_t\defeq\log \hat{\boldsymbol{\sigma}}_t^2$ to enable numerical stability. The trajectory point $\mathbf{p}_t$ follows a predictive Gaussian distribution, i.e., $\mathbf{p}_t\sim\mathcal{N}(\hat{\mathbf{p}_t},\hat{\boldsymbol{\sigma}}_t)$. As the $\mathbf{o}_{t-1}^{(\mathcal{T})}$ encodes the observation from $\mathbf{p}_{t-1}$ and its global historical context, our emission model is thus more powerful to predict $\mathbf{p}_t$. This predictive mode has also been successful in traditional methods such as the SRNN~\cite{SRNN_NIPS2016} and VRNN~\cite{VRNN_NIPS2015}. 

\vspace{-2mm}
\paragraph{Discussion}Compared to ProTran~\cite{ProTran_NIPS2021}, our formulation could individually formulate the trajectory and visual context $\mathbf{p}_{1:C}$ and $\mathbf{I}_{1:C}$ in state transition by modality-specific embeddings and the predictive link from $\mathbf{p}_{t-1}$ to $\mathbf{p}_t$, while ProTran only handles single modality context $\mathbf{p}_{1:C}$ in state transition and emission. In addition,
to learn model parameters $\theta$ and $\phi$, ProTran has to use variational posterior distribution $q_{\theta}(\mathbf{z}_t|\mathbf{z}_{1:t-1}, \mathbf{p}_{1:C})$ to help approximate Eq.~\eqref{eq:likelihood} by ELBO maximization. 
In contrast, our method does not need approximation and formulates the data uncertainty of $\mathbf{p}_t$ from a Bayesian perspective (Eq.~\eqref{eq:transition}), which is empirically more effective to handle data noise.

\subsection{Model Training}

\paragraph{Depth Robust Aleatoric Uncertainty} With the predicted trajectory points $\hat{\mathbf{p}}_t$ along with the uncertainty $\hat{\boldsymbol{\alpha}}$, according to~\cite{Unct_NIPS2017,bao2020uncertainty}, the model training is essentially to learn the heteroscedastic aleatoric uncertainty (HAU) from data. As shown in~\cite{kendall2018multi,he2019bounding}, by minimizing the KL divergence between the predictive Gaussian distributions and Dirac distribution of the ground truth trajectory, the objective in Eq.~\eqref{eq:obj} is equivalent to minimizing the HAU loss at each time $t$:
\begin{equation}
    \mathcal{L}_{\text{haul}}(\hat{\alpha}, \hat{\mathbf{p}}, \mathbf{p}) = e^{-\hat{\alpha}} \sum_{i=1}^{|\mathbf{p}|} \lVert p_{i} - \hat{p}_{i}\rVert^2+ \hat{\alpha},
\label{eq:haul}
\end{equation}
where $p_i$ and $\hat{p}_i$ are 3D coordinate values $(x,y,z)$ of ground truth and model prediction, respectively.

In our task, the trajectory depth $z$ is more challenging to predict than $x$ and $y$ due to 1) the weak implicit correspondence between the past visual context $\mathcal{V}_{1:C}$ and future hand depth, and 2) more importantly, the inevitable annotation noise from depth sensors. 
To handle these challenges, we propose to decouple the aleatoric uncertainty into $\hat{\alpha}_{t}$ which is isotropic for $(x,y)$ and $\hat{\beta}_{t}$ specifically for $z$, respectively. Then, the predictions of $(x,y)$ and $z$ are weighted by factors $w_t$ so that the regression loss becomes
\begin{equation}
    \mathcal{L}_{\text{DRAU}}^{(t)}(\hat{\mathbf{y}},\hat{\boldsymbol{\alpha}}) = \mathcal{L}_{\text{haul}}(\hat{\alpha}_{t}, \hat{\mathbf{p}}_{t}^{(\text{2d})}, \mathbf{p}_{t}^{(\text{2d})}) + w_{t} \mathcal{L}_{\text{haul}}(\hat{\beta}_{t}, \hat{z}_{t}, z_{t}),
\end{equation}
where the weight $w_t$ is determined by the negative temporal difference of ground truth $z_{1:T}$ with softmax normalization:
\begin{equation}
    w_{t} = \frac{\exp(-\Delta z_t)}{\sum_{t=1}^{T}\exp(-\Delta z_t)}, \;\;\; \Delta z_t = |z_t - z_{t-1}|.
\label{eq:depth_weight}
\end{equation}
Since $\Delta z_t$ indicates the stability of depth transition, the motivation of $w_t$ is to encourage large weight on the stable depth transitions (small $\Delta z_t$) in a trajectory, which enables the training to focus less on the unstable depth so that the model is robust to noisy depth annotations.

\paragraph{Velocity Constraints} To explicitly inject the physical rule of hand motion into the model, we additionally take the motion inertia into consideration. Specifically, we leverage the transitioned states $\{\mathbf{z}_1,\ldots,\mathbf{z}_T\}$ learned from Eq.~\eqref{eq:transition} to predict the velocities $\{\mathbf{v}_1,\ldots,\mathbf{v}_T\}$ by an MLP. Then, we propose the following velocity constraint in training:
\begin{equation}
\begin{split}
    \mathcal{L}_{\text{velo}}(\hat{\mathbf{v}},\mathbf{p}) & = \sum_{t=1}^T \left(\lVert \mathbf{p}_t - \mathbf{p}_{t-1} - \hat{\mathbf{v}}_t\rVert^2\right) \\
    &+ \gamma \sum_{t=C+1}^{T}\left(\lVert \mathbf{p}_{C} + \sum_{i=C+1}^{t} \hat{\mathbf{v}}_i - \hat{\mathbf{p}}_t\rVert^2 \right),
\label{eq:lossvelo}
\end{split}
\end{equation}
where the first term uses the first-order difference of locations $\mathbf{p}_t$ to supervise the predicted velocity $\hat{\mathbf{v}}$ and we set $\mathbf{p}_0$ to zero. The second term is to constrain the future predicted trajectory point $\hat{\mathbf{p}}_t$ with the warped point, which is computed by adding the accumulative predicted velocities onto the last observed point $\mathbf{p}_C$ since the time interval is one. Empirically, we found the velocity constraint enables better generalization capability to unseen data (see Table~\ref{tab:arch}).

\paragraph{Visual Prompt Tuning}
Visual prompt tuning (VPT)~\cite{VPT_ECCV2022} has been recently successful to adapt large visual foundation models to downstream vision tasks. In this paper, we leverage VPT to adapt pre-trained visual backbone $f_\mathcal{V}$ for the trajectory prediction task. The basic idea is to append learnable prompt embeddings $\textbf{P}$ to the input image $\textbf{I}$ and only learn a few layers of MLP head $h_{\psi}$ while keeping the backbone parameters $\Psi$ frozen as  $\Psi^*$ in training:
\begin{equation}
    \mathbf{x}_t^{(\mathcal{V})} = h_{\psi}( f_{\mathcal{V}}^{\textcolor{black}{\Psi^*}}  (\mathbf{I}_t, \textcolor{black}{\mathbf{P}} )) 
\end{equation}
where only the head parameters $\psi$ and visual prompt $\mathbf{P}$ are learned in training. Since $\{\psi,\mathbf{P}\}$ are much smaller than $\Psi$, the VPT is highly efficient in training. We implemented $f_{\mathcal{V}}$ with both ResNet~\cite{ResNet_CVPR2016} and ViT~\cite{ViT_ICLR2020}, without noting significant performance difference. However, applying VPT achieves better 3D hand trajectory prediction performance than traditional fine-tuning (see Fig.~\ref{fig:prompt_len}). This is interesting as there is no existing literature that explores the VPT for vision-based trajectory prediction problems. 

\begin{figure}[t]
    \centering
    \includegraphics[width=\linewidth]{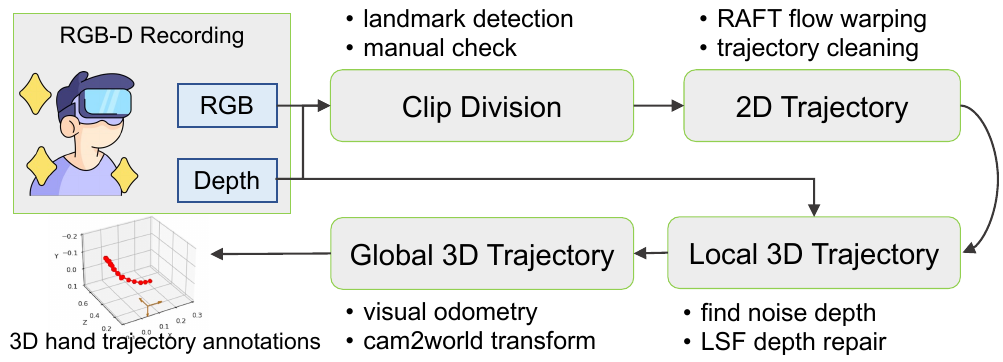}
    \caption{\small\textbf{Annotation Workflow.} For more details of the annotation procedure, please refer to our supplementary materials.}
    \label{fig:annoflow}
\end{figure}

\section{Experiments}
\label{sec:exp}

\subsection{Datasets}

Since there is no available egocentric 3D hand trajectory dataset, we collect annotations based on two existing datasets, i.e., H2O~\cite{H2O_ICCV21} and EgoPAT3D~\cite{EgoPAT3D_CVPR2022}, which contain egocentric RGB-D raw recordings for annotation purpose.

\textbf{H2O~\cite{H2O_ICCV21}} dataset is initially collected for 3D hand pose and interaction recognition using RGB-D data from both egocentric and multiple third-person views. We first use the precisely annotated 3D hand poses to compute the 3D centroids as the ground truth of the 3D hand trajectory, named \textbf{H2O-PT}, which is guaranteed to be of high-quality in~\cite{H2O_ICCV21} by multi-view verification. 

\textbf{EgoPAT3D}~\cite{EgoPAT3D_CVPR2022} dataset is much larger than H2O. It is initially collected for predicting the 3D action targets from egocentric 3D videos. However, it does not provide either the 3D hand trajectory or the 3D hand poses. Thus, similar to~\cite{EgoPAT3D_CVPR2022}, we develop an annotation workflow as shown in Fig.~\ref{fig:annoflow}. More details about the annotation workflow are in the supplement. Eventually, we obtained sufficiently large collections of 3D hand trajectories for training and evaluation, named \textbf{EgoPAT3D-DT}. To verify the reliability of the annotation workflow, we also apply it to H2O, resulting in a \textbf{H2O-DT} dataset.

\vspace{-2mm}
\paragraph{Dataset Split}
The H2O(-PT/DT) dataset consists of 184 untrimmed long videos. We temporally sample the videos into multiple 64-frame clips with a step-size of 15 frames, resulting in 8203, 1735, and 3715 samples in training, validation, and testing splits, respectively.
The EgoPAT3D-DT consists of 14 scenes and we split it into 11 seen scenes containing 8807 samples and 3 unseen testing scenes containing 2334 samples. The unseen scenes are not used in training, and the seen scenes are split into 6356, 846, and 1605 samples for training, validation, and seen testing. 

\vspace{-4mm}
\paragraph{Evaluation Setting}
We use the 3D Average Displacement Error (ADE) and Final Displacement Error (FDE) in meters as the evaluation metrics. The 2D trajectory results are normalized with reference to the video frame size. For all metrics, a small value indicates better performance. Each model is trained with 3D and 2D trajectory targets individually and evaluated by 3D metrics ($\text{3D}_{\textcolor{red}{(3D)}}$) and 2D metrics ($\text{2D}_{\textcolor{blue}{(2D)}}$), respectively. The 3D model is additionally evaluated by 2D metrics ($\text{2D}_{\textcolor{red}{(3D)}}$) by projecting the 3D trajectory outputs to the 2D image plane.

\begin{table}[t]
\centering
\captionsetup{font=small,aboveskip=3pt}
\caption{\textbf{ADE and FDE results on H2O-PT dataset}. All models are built with ResNet-18 backbone. Best and secondary results are viewed in bold \textbf{black} and \textcolor{blue}{\textbf{blue}} colors, respectively.
}
\label{tab:h2o}
\small
\setlength{\tabcolsep}{0.6mm}
\setlength{\extrarowheight}{0.5mm}
    \begin{tabular}{l|ccc|ccc}
    \toprule
    \multicolumn{1}{c|}{\multirow{2}{*}{Model}} & \multicolumn{3}{c|}{ADE ($\downarrow$)} & \multicolumn{3}{c}{FDE ($\downarrow$)} \\
\cline{2-7}          & $\text{3D}_{(\textcolor{red}{3D})}$ & $\text{2D}_{(\textcolor{red}{3D})}$  & $\text{2D}_{(\textcolor{blue}{2D})}$  & $\text{3D}_{(\textcolor{red}{3D})}$ & $\text{2D}_{(\textcolor{red}{3D})}$  & $\text{2D}_{(\textcolor{blue}{2D})}$\\
    \hline
    DKF~\cite{DKF_arXiv2015}   & 0.159  & 0.186  & 0.211  & 0.137  & 0.163  & 0.185   \\
    RVAE~\cite{RVAE_ICASSP2020} & 0.046  & 0.055  & 0.056  & 0.067  & 0.081  & \textbf{0.037}   \\
    DSAE~\cite{DSAE_ICML2018} & 0.051  & 0.060  & 0.059  & 0.057  & 0.067  & 0.076   \\
    STORN~\cite{STORN_NIPSW2014} & 0.043  & 0.053  & 0.053  & 0.094  & 0.141  &  0.076  \\
    VRNN~\cite{VRNN_NIPS2015} & 0.041  & 0.050  & 0.050  &  0.051 & 0.081  & 0.068   \\
    SRNN~\cite{SRNN_NIPS2016} & 0.040  & 0.048  & 0.049  & \textcolor{blue}{0.036}  & \textcolor{blue}{0.061}  & 0.044   \\
    EgoPAT3D\footnotemark[1]~\cite{EgoPAT3D_CVPR2022}   & 0.039  & 0.046  & \textcolor{blue}{0.048}  & \textbf{0.034}  & 0.064  & 0.060   \\
    \hline
    AGF\footnotemark[1]~\cite{AGF_ICCV21} & \textcolor{blue}{0.039}  & \textcolor{blue}{0.046}  &  0.081  & 0.069  & 0.065  &  0.146  \\  
    OCT\footnotemark[1]~\cite{OCT_CVPR2022}   & 0.252  & 0.311  & 0.387  & 0.278  & 0.471  & 0.381   \\
    ProTran\footnotemark[1]~\cite{ProTran_NIPS2021} & 0.066  & 0.088  & 0.109  & 0.099  & 0.168  &  0.123  \\
    \hline
    USST & \textbf{0.031}  & \textbf{0.037}  & \textbf{0.040}  & 0.052  & \textbf{0.043}  &  \textcolor{blue}{0.043}  \\
    \bottomrule
    \end{tabular}%
\end{table}%

\subsection{Implementation Detail}
The proposed method is implemented by PyTorch. In pre-processing, RGB videos are down-scaled to $64\times 64$. The 3D global trajectory data are normalized and further centralized to the range [-1,1]. By default, we set the observation ratio to 60\%, the feature dimensions of $\mathbf{o}^{(\mathcal{V})}$ and $\mathbf{o}^{(\mathcal{T})}$ to 256, and the dimension of $\mathbf{z}$ to 16 for all methods. In training, we use Huber loss to compute the location error. We adopt the Adam optimizer with base learning rate 1e-4 and cosine warmup scheduler for 500 training epochs on EgoPAT3D and 350 epochs on H2O datasets, respectively. More implementation details are in the supplement.

\begin{table}[t]
\centering
\captionsetup{font=small,aboveskip=3pt}
\caption{\textbf{ADE results on EgoPAT3D-DT dataset}. All models are built with ResNet-18 backbone. Best and secondary results are viewed in bold \textbf{black} and \textcolor{blue}{\textbf{blue}} colors, respectively.}
\label{tab:main}
\small
\setlength{\tabcolsep}{0.6mm}
\setlength{\extrarowheight}{0.5mm}
    \begin{tabular}{l|ccc|ccc}
    \toprule
    \multicolumn{1}{c|}{\multirow{2}{*}{Model}} & \multicolumn{3}{c|}{Seen Scenes ($\downarrow$)} & \multicolumn{3}{c}{Unseen Scenes ($\downarrow$)} \\
\cline{2-7}          & $\text{3D}_{(\textcolor{red}{3D})}$ & $\text{2D}_{(\textcolor{red}{3D})}$  & $\text{2D}_{(\textcolor{blue}{2D})}$  & $\text{3D}_{(\textcolor{red}{3D})}$ & $\text{2D}_{(\textcolor{red}{3D})}$  & $\text{2D}_{(\textcolor{blue}{2D})}$\\
    \hline
    DKF~\cite{DKF_arXiv2015}   &   0.294   & 0.237 &   0.157    &   0.260   & 0.202 &  0.133 \\
    RVAE~\cite{RVAE_ICASSP2020} &  0.216  & 0.110  &  0.121     &  0.194   & 0.104 &  0.109 \\
    DSAE~\cite{DSAE_ICML2018} &  0.214 &  0.129  &   0.143    &   0.188  & 0.116 &  0.131 \\
    STORN~\cite{STORN_NIPSW2014} &  0.194   & 0.092 &   0.083    &  \textcolor{blue}{0.161} &  0.084  & 0.070 \\
    VRNN~\cite{VRNN_NIPS2015} &   0.194  & 0.092 &   0.083    &  0.164   & 0.086 &  0.070 \\
    SRNN~\cite{SRNN_NIPS2016} &  0.192   & \textcolor{blue}{0.088} &  \textcolor{blue}{0.079}     &  0.166 &  0.081  & \textcolor{blue}{0.067} \\
    EgoPAT3D\footnotemark[1]~\cite{EgoPAT3D_CVPR2022}  &   \textcolor{blue}{0.186}    &  \textbf{0.081}  &   \textbf{0.079}     &  0.170    & \textcolor{blue}{0.080}   &  0.068  \\
    \hline
    AGF\footnotemark[1]~\cite{AGF_ICCV21} & 6.149  & 0.136  & 0.099  & 6.045  & 0.119  & 0.087 \\
    OCT\footnotemark[1]~\cite{OCT_CVPR2022}   &  0.853  & 0.163  &  0.098    &   0.782 & 0.139  &  0.091 \\
    ProTran\footnotemark[1]~\cite{ProTran_NIPS2021} &  0.314  & 0.179  &  0.135     &  0.240 & 0.154   & 0.107 \\
    \hline
    USST &  \textbf{0.183}  & 0.089  &  0.082   &  \textbf{0.120} & \textbf{0.075} &  \textbf{0.060} \\  
    \bottomrule
    \end{tabular}%
  \label{tab:addlabel}%
\end{table}%
\footnotetext[1]{We adapted the task-specific outputs of EgoPAT3D, AGF, OCT, and ProTran to fulfill the 3D hand trajectory forecasting task in this paper.}

\subsection{Main Results}

Table~\ref{tab:h2o} and~\ref{tab:main} show the comparison between our method and existing sequential prediction approaches on H2O-PT and EgoPAT3D-DT datasets, respectively. The methods in the first multi-row section are general RNN-based while those in the second multi-row section show recent Transformer-based models. We put the FDE results on EgoPAT3D-DT in the supplement. The tables show our method achieves the best ADE performance and comparable FDE results with AGF and SRNN on H2O-PT, and significantly outperforms AGF, OCT, and ProTran on EgoPAT3D-DT. The competitive performance of SRNN is because of its both forward and backward passes over time such that all future positional encodings are utilized for forecasting. 
We notice that AGF, OCT, and ProTran do not work well on EgoPAT3D-DT, potentially due to the KL divergence vanish issue. The higher performance on the unseen split than the seen split can be attributed to the less distribution shift between unseen test trajectories and the training trajectories.

\subsection{Model Analysis}

\begin{table}[t]
\centering
\small
\setlength{\tabcolsep}{3.0mm}
\setlength{\extrarowheight}{0.3mm}
\captionsetup{font=small,aboveskip=3pt}
  \caption{\small\textbf{Ablation Study}. All models are trained with 3D targets and tested with both 3D and 2D ADE.}
    \begin{tabular}{l|cc|cc}
    \toprule
    \multicolumn{1}{c|}{\multirow{2}{*}{Variants}} & \multicolumn{2}{c|}{Seen ($\downarrow$)} & \multicolumn{2}{c}{Unseen ($\downarrow$)} \\
\cline{2-5}          & 3D  & 2D & 3D  & 2D \\
    \hline
    ProTran (svi)~\cite{ProTran_NIPS2021} & 0.314 & 0.179  & 0.240 & 0.154 \\
    ProTran (det)~\cite{ProTran_NIPS2021}    & 0.201 &  0.104 & 0.195 &  0.106\\  
    SST (ours)  & \textbf{0.190}  & \textbf{0.088}  & \textbf{0.174}  & \textbf{0.084}  \\
    \hline
    USST w/o. $\mathcal{L}_{\text{haul}}$  & 0.292  & 0.237  &  0.267  & 0.204   \\
    USST w/o. $\mathcal{T}_{1:C}$  & 0.244   & 0.176   & 0.267   & 0.208   \\
    USST w/o. $\mathbf{p}_{t-1}$  & 0.196 & 0.090 & 0.169 & 0.098 \\
    USST w/o. $\mathcal{L}_{\text{velo}}$ & 0.189 & 0.091  & 0.168 & 0.099 \\  
    USST w/o. $w_t$ & 0.183  & 0.090  & 0.130   &  0.077   \\
    USST (full model) & \textbf{0.183}  & \textbf{0.089} & \textbf{0.120}  & \textbf{0.075} \\
    \bottomrule
    \end{tabular}%
  \label{tab:arch}%
\end{table}%

\begin{figure}
\begin{floatrow}
\ffigbox{%
    \centering
    \includegraphics[width=0.9\linewidth]{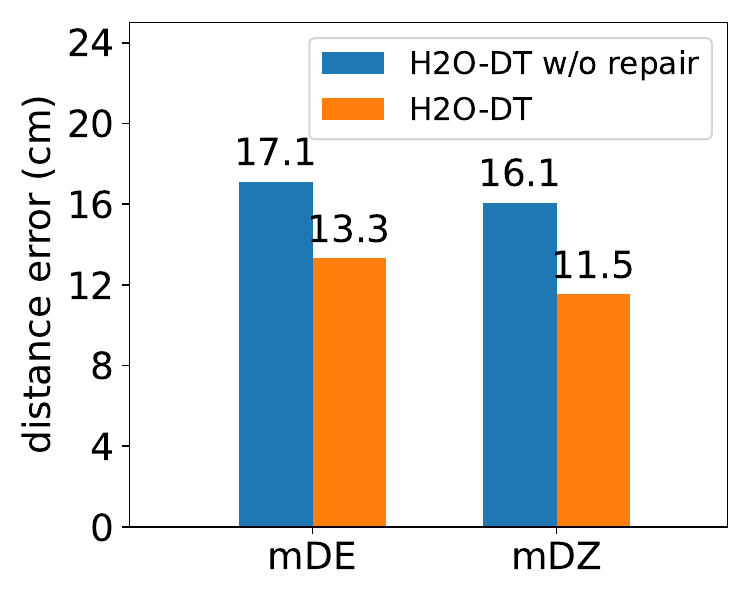}
    \label{fig:my_label}
}{\captionsetup{font=small,aboveskip=-10pt,belowskip=-10pt}
  \caption{\textbf{Effect of depth repair}. mDE and mDZ are the mean error of 3D displacement and depth, respectively.}
  \label{fig:depth_repair}}
\hspace{-0.6cm}
\capbtabbox{%
    \footnotesize
    \setlength{\tabcolsep}{0.5mm}
    \setlength{\extrarowheight}{1.2mm}
    \begin{tabular}{l|c|c|c}
    \toprule
         \multicolumn{2}{c|}{Metrics} & SRNN & USST \\
    \hline
         \parbox[t]{2mm}{\multirow{2}{*}{\rotatebox[origin=c]{90}{\scriptsize{H2O-\textbf{DT}}}}} & ADE  & .087 / .076  &  .033 / .041 \\
            & FDE & .124 / .045  & .052 / .041 \\
    \hline
         \parbox[t]{2mm}{\multirow{2}{*}{\rotatebox[origin=c]{90}{\scriptsize{H2O-\textbf{PT}}}}}  & ADE  & .040 / .049  &  .031 / .040 \\
          & FDE & .036 / .044  & .052 / .043 \\
    \bottomrule
    \end{tabular}
}{\captionsetup{font=small,aboveskip=-10pt,belowskip=-5pt}\caption{\textbf{Annotation Reliability}. ADE results ($\text{3D}_{(\textcolor{red}{3D})}$ / $\text{2D}_{(\textcolor{blue}{2D})}$) are from testing on the same H2O-PT test set.}
\label{tab:reliability}}
\end{floatrow}
\end{figure}

\paragraph{Ablation Study} To validate the effectiveness of the proposed modules and loss functions, we report the results of the ablation study in Table~\ref{tab:arch} on the EgoPAT3D-DT dataset.

We first compare the ProTran with the proposed state space transformer (SST), which is a vanilla version of USST without uncertainty modeling, velocity constraint, and VPT. For a fair comparison, we implement a deterministic (det) version of ProTran in addition to the original method that uses stochastic variational inference (svi). Table~\ref{tab:arch} shows a clear advantage of our method over ProTran, which demonstrates the superiority of our SSM for state transition.

Next, in Table~\ref{tab:arch}, we individually remove the new components and compare them with the full model of USST, including 1) the uncertainty loss function $\mathcal{L}_{\text{haul}}$, 2) the trajectory context $\mathcal{T}_{1:C}$, 3) the predictive link in $p_{\phi}(\mathbf{p}_t|\mathbf{z}_t,\mathbf{p}_{t-1})$, 4) the velocity constraint $\mathcal{L}_{\text{velo}}$, and 5) the depth robust weight $w_t$. It shows that uncertainty modeling is critical to guarantee reasonable forecasting results. Without historical trajectory $\mathcal{T}_{1:C}$, as expected, the performance degradation is significant. The predictive link from $\mathbf{p}_{t-1}$ to $\mathbf{p}_t$ is also important for the forecasting problem, which is consistent with the recent finding in~\cite{girin2021dynamical}. It is noticeable that the velocity constraint shows a larger performance gain (4.8cm of 3D trajectory) on the unseen test set than on the seen data (0.6cm of 3D trajectory), revealing the importance of the physical rule for generalizable trajectory prediction. Lastly, the depth robust weight $w_t$ (Eq.~\eqref{eq:depth_weight}) also boosts the performance of unseen data, showing the importance of modeling the depth noise from annotations.

\begin{figure}[t]
    \centering
    \includegraphics[width=0.47\linewidth]{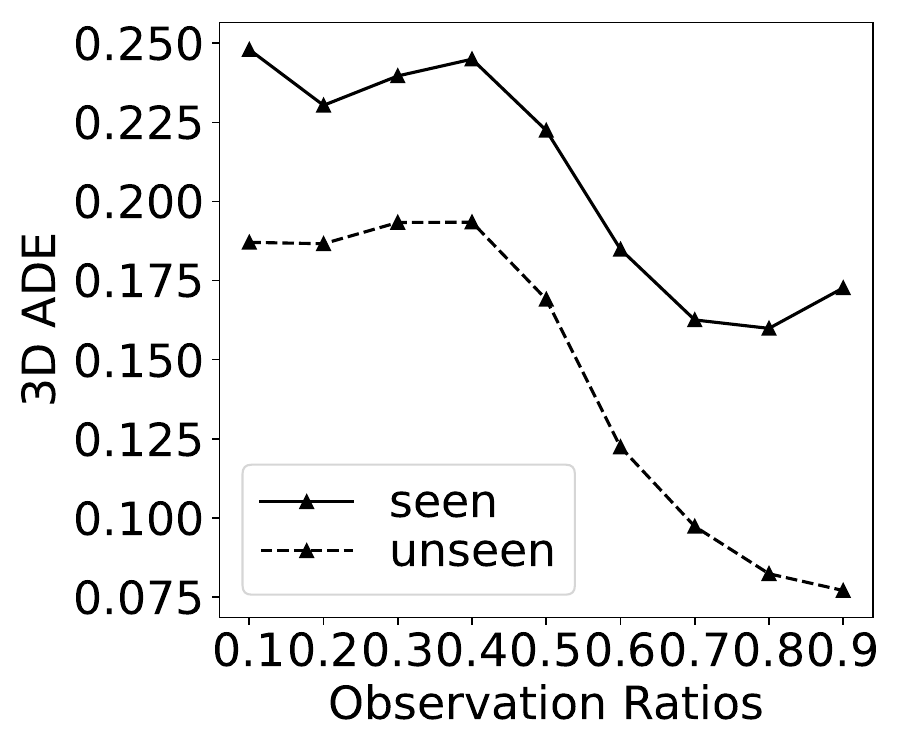}
    \includegraphics[width=0.47\linewidth]{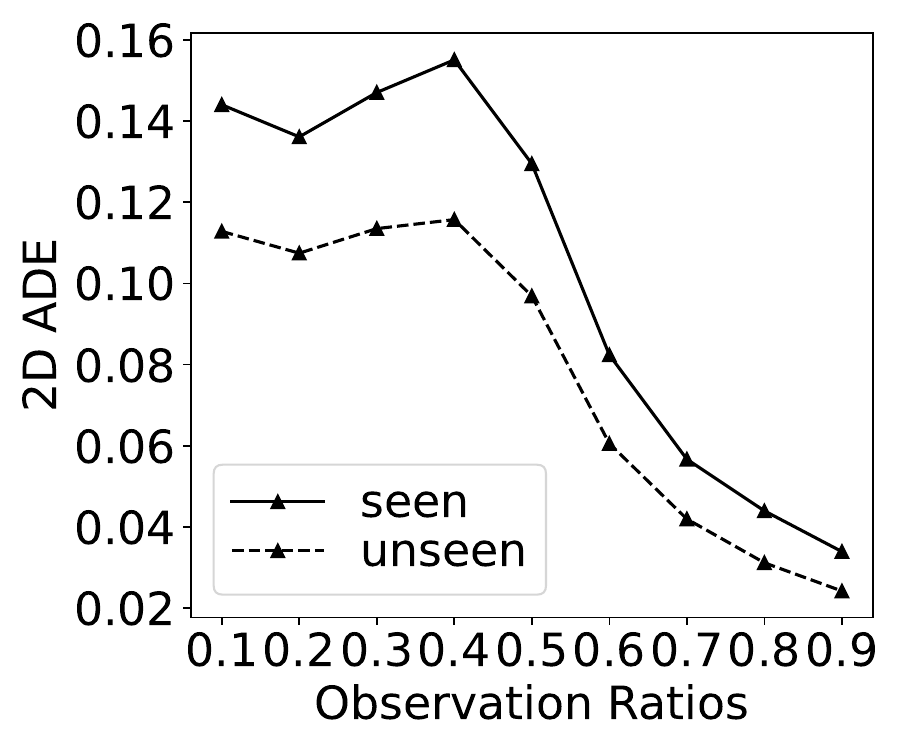}
    \captionsetup{font=small,aboveskip=5pt,belowskip=-5pt}
    \caption{\textbf{Arbitrary Observation Ratios.} We report the results of 3D ADE (left) and 2D ADE (right) on EgoPAT3D-DT dataset.}
    \label{fig:ratios}
\end{figure}

\begin{figure}[t]
    \centering
    \includegraphics[width=0.47\linewidth]{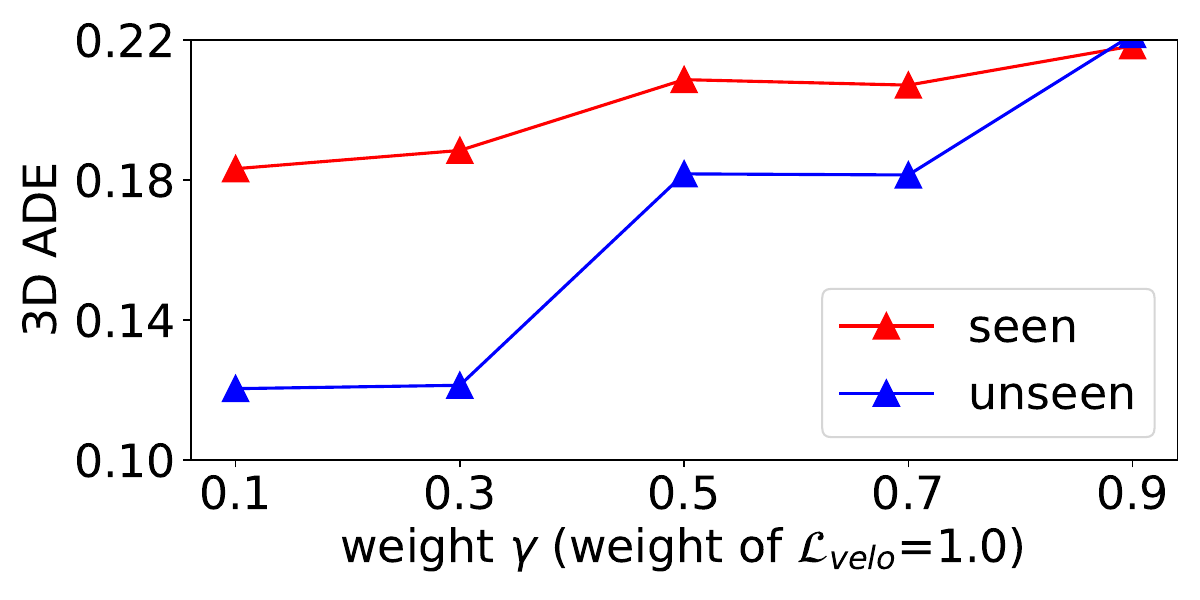}
    \includegraphics[width=0.47\linewidth]{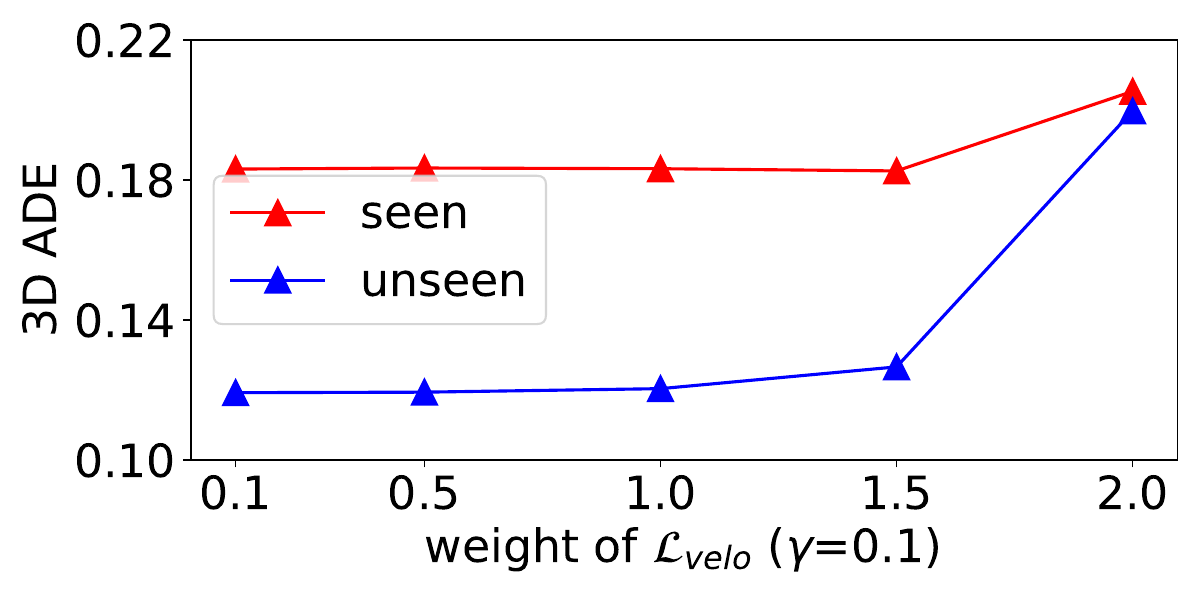}
    \captionsetup{font=small,aboveskip=5pt,belowskip=-10pt}
    \caption{\small{\textbf{Impact of loss weights.} Left: set the weight of $\mathcal{L}_{\text{velo}}$ to 1.0 and tune $\gamma$. Right: set $\gamma$ to 0.1 and tune the weight of $\mathcal{L}_{\text{velo}}$.}}
    \label{fig:lossw}
\end{figure}

\begin{figure}[t]
    \centering
    \includegraphics[width=0.47\linewidth]{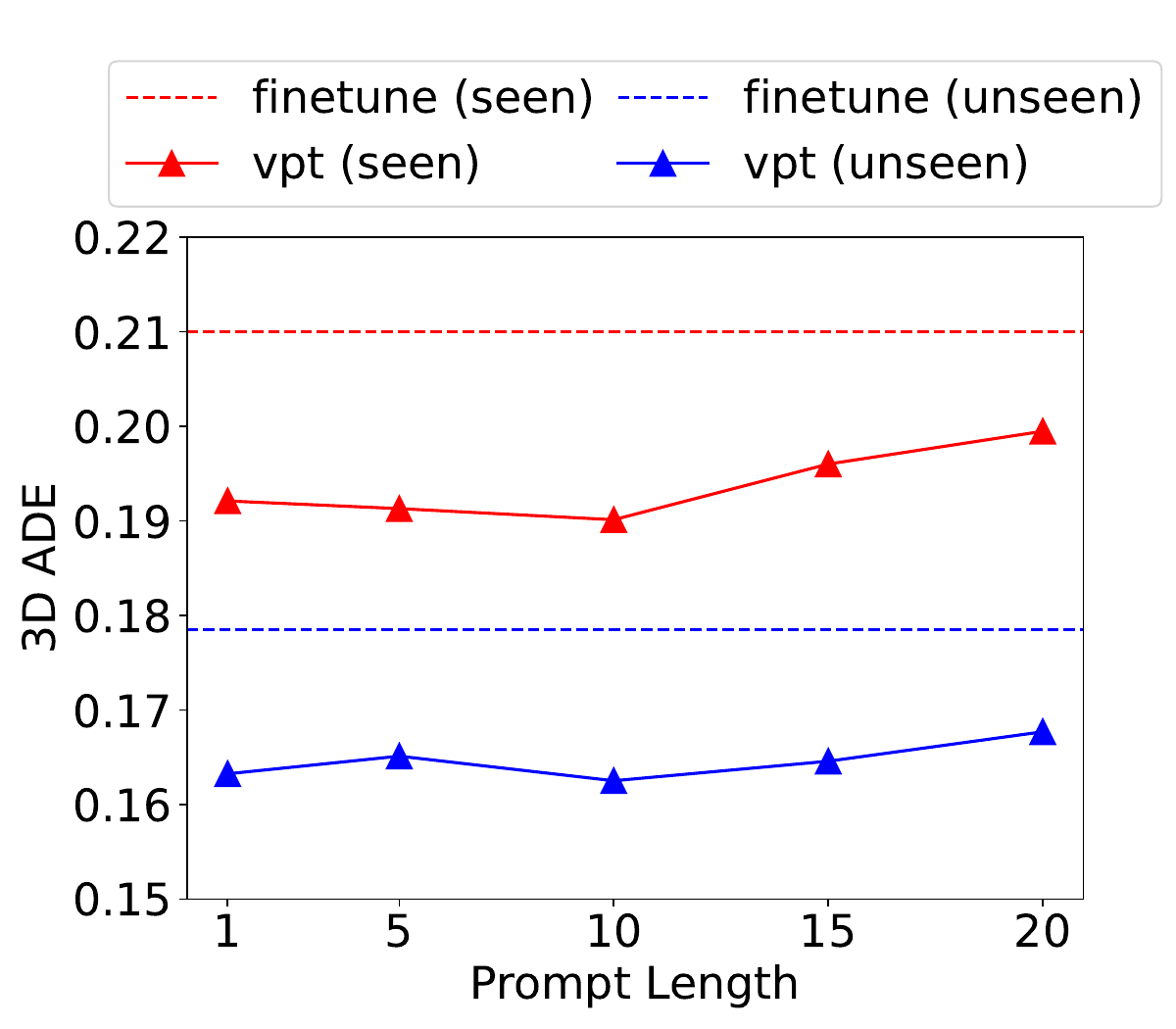}
    \includegraphics[width=0.47\linewidth]{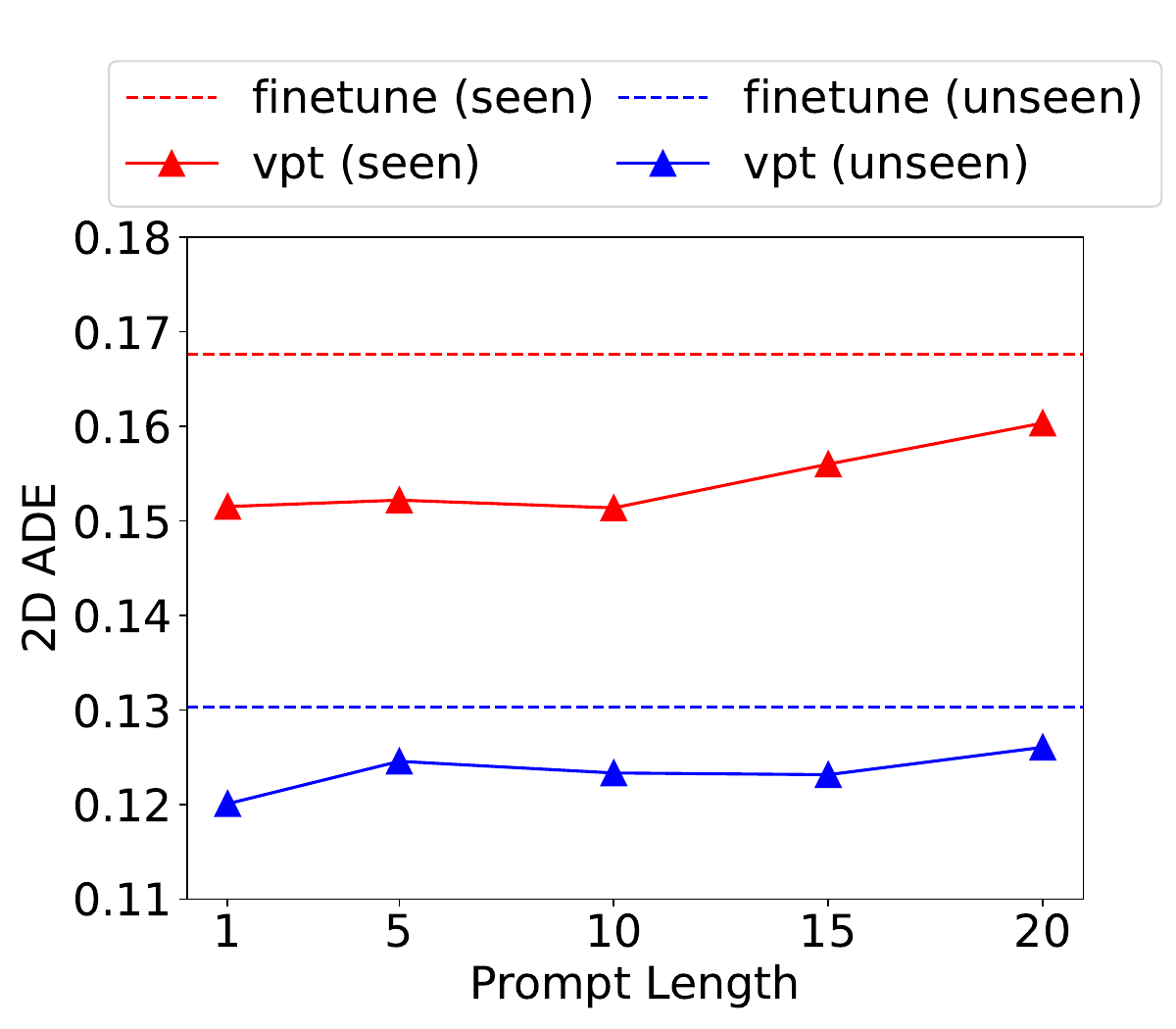}
    \captionsetup{font=small,aboveskip=5pt,belowskip=-5pt}
    \caption{\textbf{Impact of Prompt Length.} We report the results of 3D (left) and 2D (right) ADE on EgoPAT3D-DT. The ``finetune (unseen)" means finetune model on seen but test on unseen scenes.
    }
    \label{fig:prompt_len}
\end{figure}

\begin{figure*}[t]
\footnotesize
\centering
\renewcommand{\tabcolsep}{0.7pt} %
\begin{tabular}{ccccc}
& Microwave (\textbf{seen}) & PantryShelf (\textbf{seen}) & StoveTop (\textbf{unseen}) & Windowsill (\textbf{unseen})
\\
\parbox[c]{4mm}{\multirow{1}{*}[5.5em]{\rotatebox[origin=c]{90}{SRNN~\cite{SRNN_NIPS2016}}}} &
\includegraphics[width=\figwidth]{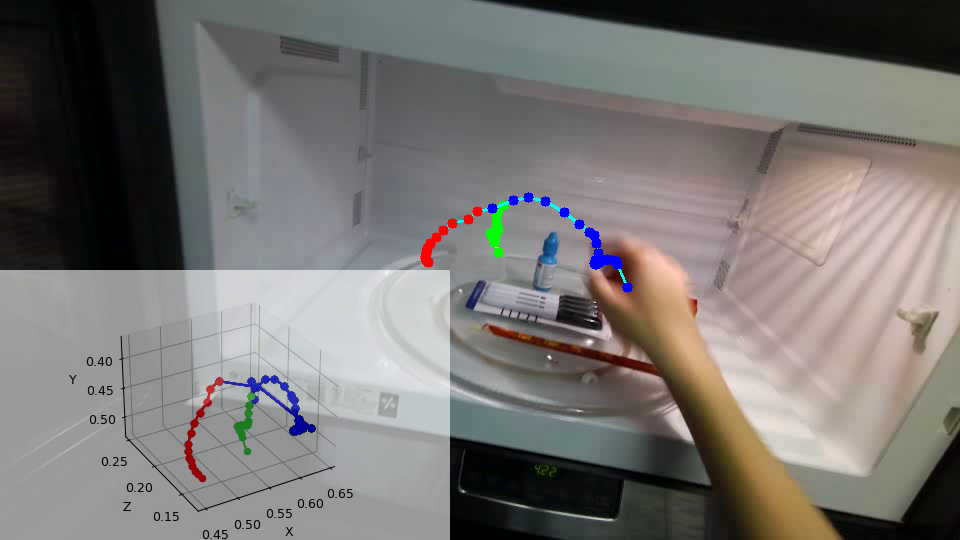} &
\includegraphics[width=\figwidth]{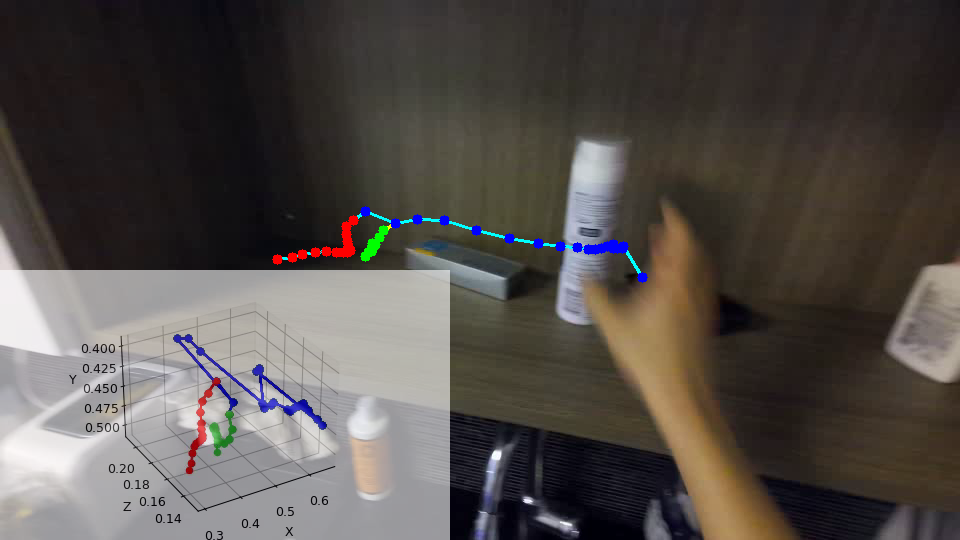} &
\includegraphics[width=\figwidth]{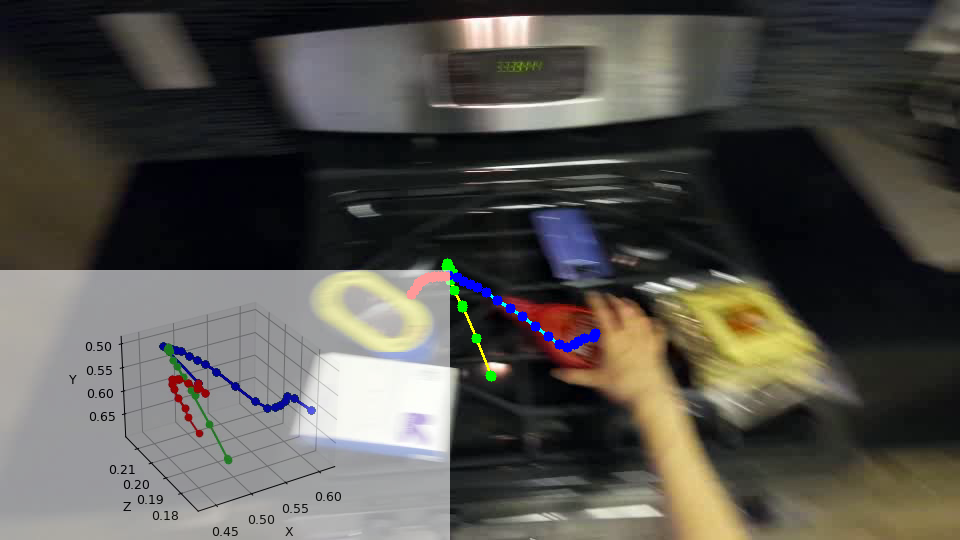} &
\includegraphics[width=\figwidth]{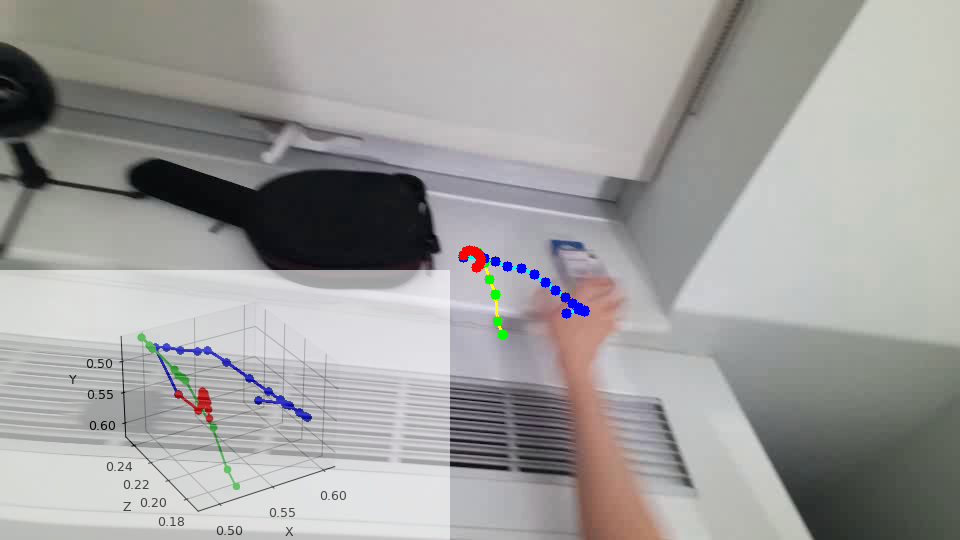}
\\
\parbox[c]{4mm}{\multirow{1}{*}[5.5em]{\rotatebox[origin=c]{90}{ProTran~\cite{ProTran_NIPS2021}}}} &
\includegraphics[width=\figwidth]{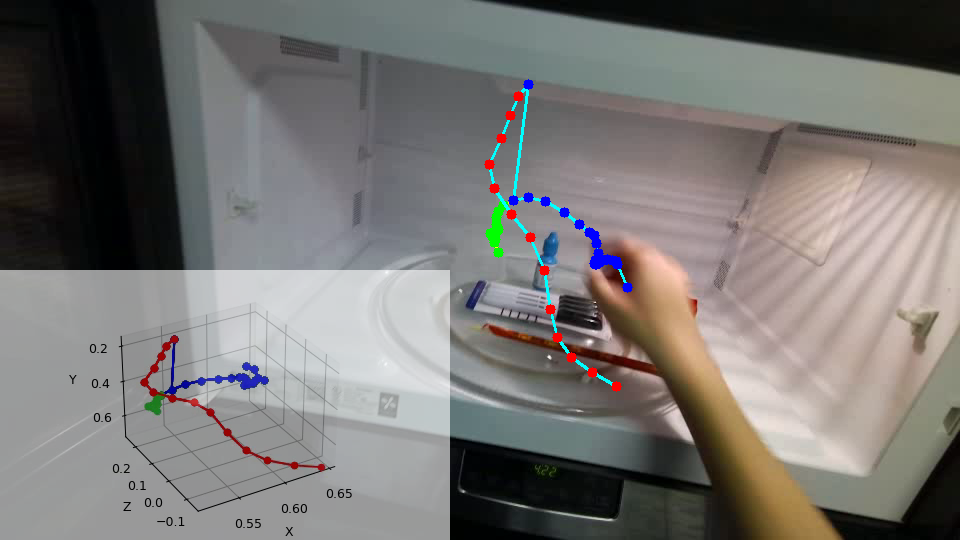} &
\includegraphics[width=\figwidth]{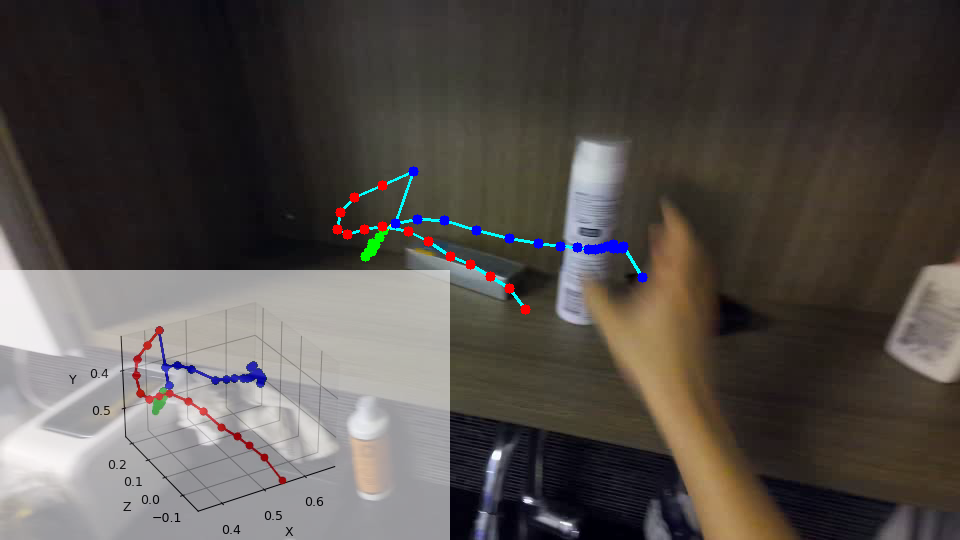} &
\includegraphics[width=\figwidth]{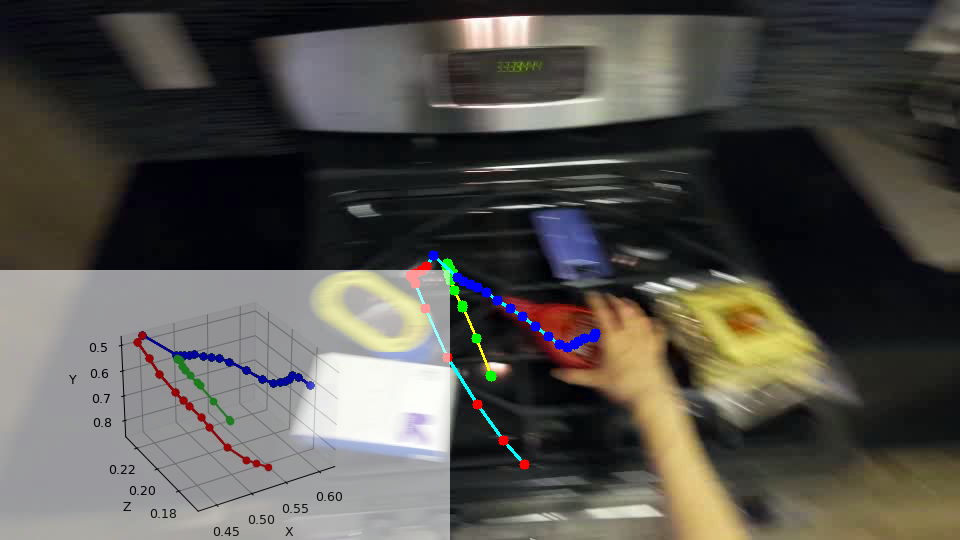} &
\includegraphics[width=\figwidth]{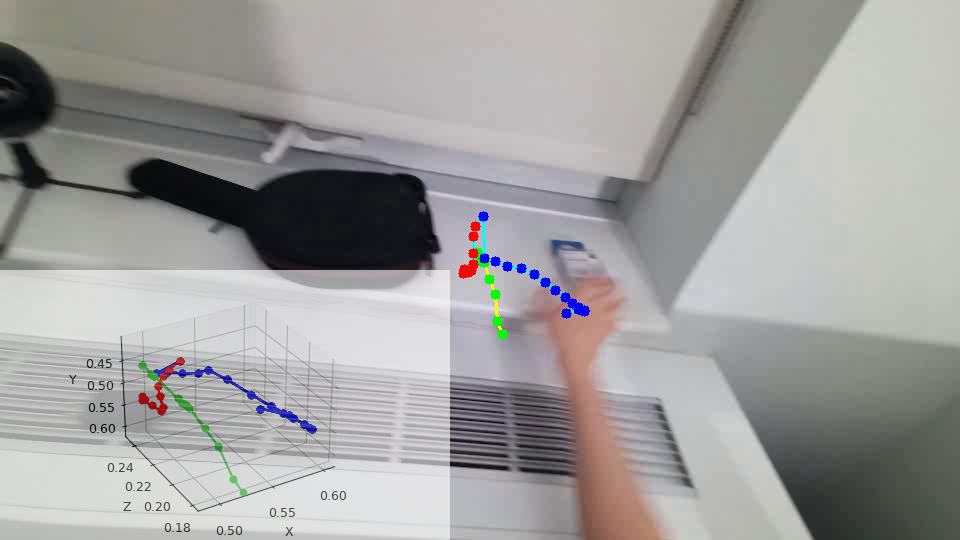}
\\
\parbox[t]{4mm}{\multirow{1}{*}[5.5em]{\rotatebox[origin=c]{90}{USST (Ours)}}} &
\includegraphics[width=\figwidth]{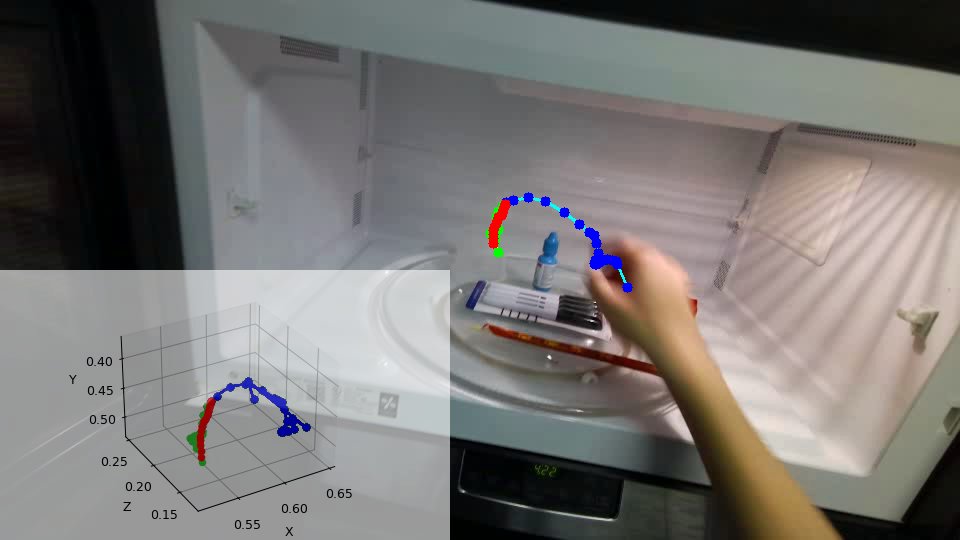} &
\includegraphics[width=\figwidth]{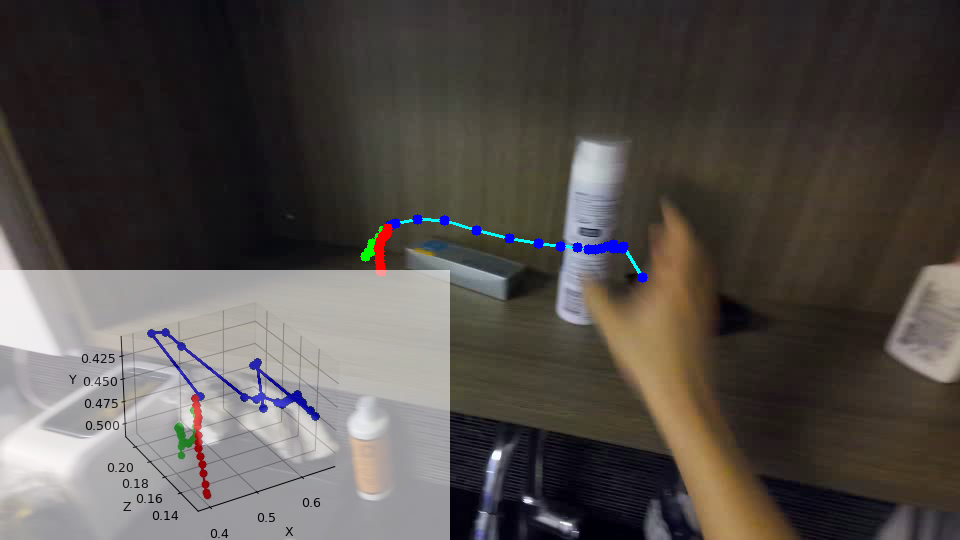} &
\includegraphics[width=\figwidth]{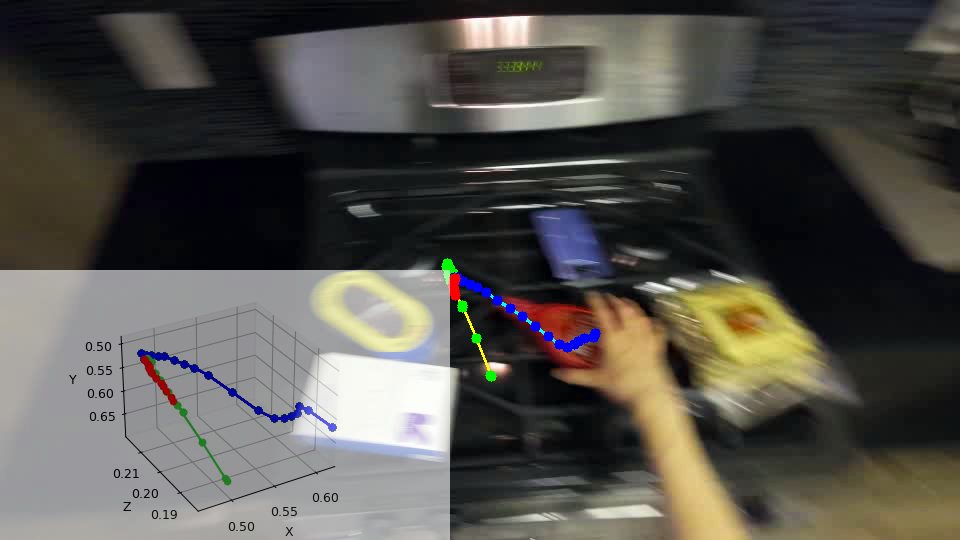} &
\includegraphics[width=\figwidth]{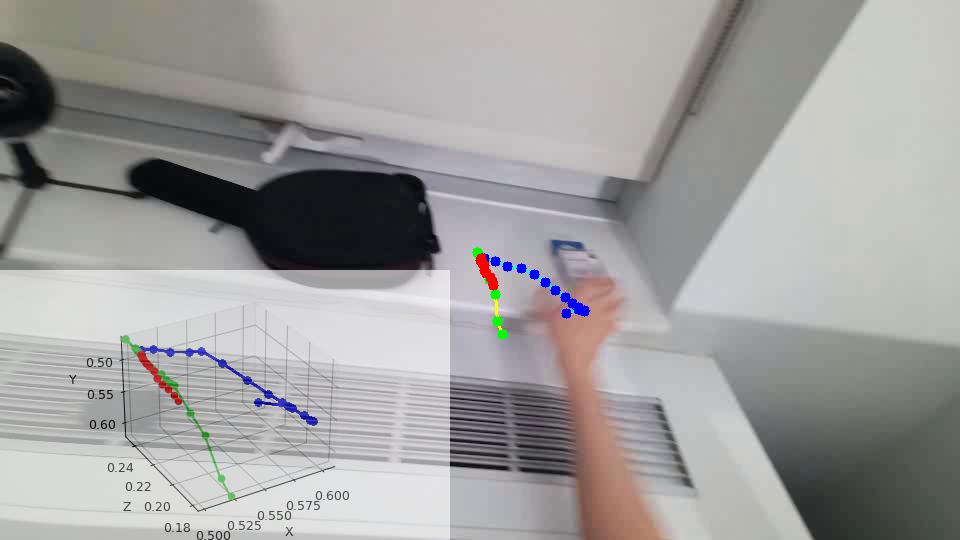}
\\
\end{tabular}
\captionsetup{font=small,aboveskip=3pt}
\caption{\textbf{Visualization on EgoPAT3D.} For each example (in a column), we show the global 3D trajectory and its 2D projection on the first frame. The \textcolor{blue}{blue}, \textcolor{green}{green}, and \textcolor{red}{red} trajectory points represent the past observed, future ground truth, and future predictions, respectively. 
}
\label{fig:vis}
\vspace{-10pt}
\end{figure*}

\vspace{-2mm}
\paragraph{Annotation Reliability}By using the accurate H2O-PT as a reference, in Fig.~\ref{fig:depth_repair}, we show the effect of repairing depth of the 3D trajectory annotations from raw RGB-D data. We see a clear improvement in mDE and mDZ measurements. In Table~\ref{tab:reliability}, we further show the performance impact of annotation quality on SRNN and our USST models. It shows that our USST achieves more consistent ADE and FDE results than SRNN over the H2O-PT and H2O-DT. 
For reference, in the supplement, we additionally report the full results and analysis on H2O-DT and H2O-DT w/o repair.

\vspace{-3mm}
\paragraph{Forecast at Any Time} To simulate the real-world practice that forecasting trajectory at an arbitrary time, we take the advantage of the Transformer attention mask to fulfill random observation ratios ranging from 10\% to 90\%. The results are summarized in Fig.~\ref{fig:ratios}. It shows that with more percentage of information observed, both the 2D and 3D forecasting error are reduced as expected. It is interesting to see the slight increase of 3D ADE for the seen test data when using more observations. It could be caused by more inaccurate trajectory depth values at the end of trajectories.

\vspace{-2mm}
\paragraph{Loss Weights} Fig.~\ref{fig:lossw} shows the EgoPAT3D-DT results of tuning the weights in Eq.~\eqref{eq:lossvelo}, where the best performance is achieved when $\gamma$ is set to 0.1 and the weight of $\mathcal{L}_{\text{velo}}$ is set to 1.0, respectively. We apply them to H2O-PT by default.

\vspace{-2mm}
\paragraph{Prompt Length of VPT} As indicated in VPT literature~\cite{VPT_ECCV2022}, the length of the visual prompt in ViT models needs to be carefully tuned for downstream tasks. In experiments, based on the SST model, we select the prompt length from $\{1,5,10,15,20\}$ and compare their performance with the baseline that fine-tunes the entire ViT backbone. Results are reported in Fig.~\ref{fig:prompt_len}. It shows that VPT could steadily achieve lower 2D and 3D ADE than fine-tuning, and the best performance is achieved when the prompt length is 10.

\begin{table}[t]
\centering
\small
\setlength{\tabcolsep}{2.0mm}
\setlength{\extrarowheight}{0.3mm}
\captionsetup{font=small,aboveskip=3pt,belowskip=2pt}
  \caption{\small\textbf{Impact of 3D coordinate systems.} ``Local" and ``global" mean using 3D camera and world coordinates, respectively.}
    \begin{tabular}{l|c|cc|cc}
    \toprule
    \multicolumn{1}{c|}{\multirow{2}{*}{3D Target}} & \multicolumn{1}{c|}{\multirow{2}{*}{Backbone}}  & \multicolumn{2}{c|}{Seen ($\downarrow$)} & \multicolumn{2}{c}{Unseen ($\downarrow$)} \\
\cline{3-6}  & & \multicolumn{1}{c}{3D}  & \multicolumn{1}{c|}{2D}  & \multicolumn{1}{c}{3D} & \multicolumn{1}{c}{2D} \\
    \hline
    Local & R18 & 0.202   & \textbf{0.083}    &  0.174  & \textbf{0.062}  \\
    Global & R18 & \textbf{0.183}  & 0.089    &   \textbf{0.120}    & 0.075   \\
    \hline
    Local & ViT & 0.183    &  \textbf{0.081}   &  0.133  &  \textbf{0.067} \\
    Global & ViT & \textbf{0.182}   &  0.087 &    \textbf{0.119}   &  0.075  \\
    \bottomrule
    \end{tabular}%
  \label{tab:localglobal}%
\end{table}%

\vspace{-2mm}
\paragraph{Local vs Global 3D Trajectory} We note that the ambiguity of learning the appearance-location mapping exists when using local 3D targets. To justify the choice of global 3D trajectory targets, in Table~\ref{tab:localglobal}, we compare the 3D and 2D ADE results using both ResNet-18 and ViT as visual backbones. It clearly shows that for 3D trajectory prediction, a global 3D coordinate system is a better choice, while for 2D trajectory evaluation, the local 3D target is better. These observations are expected as in the local 3D coordinate system, the projected 2D pixel locations of moving hands tend to be in the visual center due to the egocentric view so that the model training is dominated by the 2D hand locations.

\vspace{-2mm}
\paragraph{Qualitative Results}

As shown in Fig.~\ref{fig:vis}, the proposed USST model is compared with the Transformer-based approach ProTran and the most competitive method SRNN. It clearly shows that our trajectory forecasting is much better than the three compared methods. Please refer to our supplement for more visualizations.

\vspace{-3mm}
\paragraph{Limitations \& Discussions} The dataset annotation is limited in scenarios when the RGB-D sensors or camera poses are not available. The model is limited in the recursive way of state transition, which is not hardware friendly for parallel inference. Besides, in the future, other tasks like the 3D hand pose and interaction recognition can be jointly studied for a fine-grained egocentric understanding.

\section{Conclusion}
In this paper, we propose to forecast human hand trajectory in 3D physical space from egocentric videos. For this goal, we first develop a pipeline to automate the 3D trajectory annotation. Then, we propose a novel uncertainty-aware state space transformer (USST) model to fulfill the task. 
Empirically, with the aleatoric uncertainty modeling, velocity constraint, and visual prompt tuning, our model achieves the best performance on both H2O and EgoPAT3D datasets and good generalization to the unseen scenes.


{\small
\bibliographystyle{ieee_fullname}
\bibliography{egbib}

\begin{thebibliography}{10}\itemsep=-1pt

\bibitem{alahi2016social}
Alexandre Alahi, Kratarth Goel, Vignesh Ramanathan, Alexandre Robicquet, Li
  Fei-Fei, and Silvio Savarese.
\newblock Social lstm: Human trajectory prediction in crowded spaces.
\newblock In {\em CVPR}, 2016.

\bibitem{bao2019monofenet}
Wentao Bao, Bin Xu, and Zhenzhong Chen.
\newblock Monofenet: Monocular 3d object detection with feature enhancement
  networks.
\newblock {\em IEEE TIP}, 29:2753--2765, 2019.

\bibitem{bao2020uncertainty}
Wentao Bao, Qi Yu, and Yu Kong.
\newblock Uncertainty-based traffic accident anticipation with spatio-temporal
  relational learning.
\newblock In {\em ACM MM}, 2020.

\bibitem{STORN_NIPSW2014}
Justin Bayer and Christian Osendorfer.
\newblock Learning stochastic recurrent networks.
\newblock In {\em NeurIPS Workshop}, 2014.

\bibitem{Bi_ECCV2020}
Huikun Bi, Ruisi Zhang, Tianlu Mao, Zhigang Deng, and Zhaoqi Wang.
\newblock How can i see my future? fvtraj: Using first-person view for
  pedestrian trajectory prediction.
\newblock In {\em ECCV}, 2020.

\bibitem{VRNN_NIPS2015}
Junyoung Chung, Kyle Kastner, Laurent Dinh, Kratarth Goel, Aaron~C Courville,
  and Yoshua Bengio.
\newblock A recurrent latent variable model for sequential data.
\newblock In {\em NeurIPS}, 2015.

\bibitem{EK50_ECCV2018}
Dima Damen, Hazel Doughty, Giovanni~Maria Farinella, Sanja Fidler, Antonino
  Furnari, Evangelos Kazakos, Davide Moltisanti, Jonathan Munro, Toby Perrett,
  Will Price, et~al.
\newblock Scaling egocentric vision: The epic-kitchens dataset.
\newblock In {\em ECCV}, 2018.

\bibitem{EK100_PAMI2020}
Dima Damen, Hazel Doughty, Giovanni~Maria Farinella, Sanja Fidler, Antonino
  Furnari, Evangelos Kazakos, Davide Moltisanti, Jonathan Munro, Toby Perrett,
  Will Price, et~al.
\newblock The epic-kitchens dataset: Collection, challenges and baselines.
\newblock {\em IEEE TPAMI}, 43(11):4125--4141, 2020.

\bibitem{deo2018convolutional}
Nachiket Deo and Mohan~M Trivedi.
\newblock Convolutional social pooling for vehicle trajectory prediction.
\newblock In {\em CVPR}, 2018.

\bibitem{diller2022forecasting}
Christian Diller, Thomas Funkhouser, and Angela Dai.
\newblock Forecasting characteristic 3d poses of human actions.
\newblock In {\em CVPR}, 2022.

\bibitem{ViT_ICLR2020}
Alexey Dosovitskiy, Lucas Beyer, Alexander Kolesnikov, Dirk Weissenborn,
  Xiaohua Zhai, Thomas Unterthiner, Mostafa Dehghani, Matthias Minderer, Georg
  Heigold, Sylvain Gelly, et~al.
\newblock An image is worth 16x16 words: Transformers for image recognition at
  scale.
\newblock In {\em ICLR}, 2020.

\bibitem{fathi2011understanding}
Alireza Fathi, Ali Farhadi, and James~M Rehg.
\newblock Understanding egocentric activities.
\newblock In {\em ICCV}, 2011.

\bibitem{SRNN_NIPS2016}
Marco Fraccaro, S{\o}ren~Kaae S{\o}nderby, Ulrich Paquet, and Ole Winther.
\newblock Sequential neural models with stochastic layers.
\newblock In {\em NeurIPS}, 2016.

\bibitem{Fu_NACCL2019}
Hao Fu, Chunyuan Li, Xiaodong Liu, Jianfeng Gao, Asli Celikyilmaz, and Lawrence
  Carin.
\newblock Cyclical annealing schedule: A simple approach to mitigating {KL}
  vanishing.
\newblock In {\em NACCL}, 2019.

\bibitem{gamage2021so}
Nisal~Menuka Gamage, Deepana Ishtaweera, Martin Weigel, and Anusha Withana.
\newblock So predictable! continuous 3d hand trajectory prediction in virtual
  reality.
\newblock In {\em UIST}, 2021.

\bibitem{Girdhar_ICCV2021}
Rohit Girdhar and Kristen Grauman.
\newblock Anticipative video transformer.
\newblock In {\em ICCV}, 2021.

\bibitem{girin2021dynamical}
Laurent Girin, Simon Leglaive, Xiaoyu Bie, Julien Diard, Thomas Hueber, and
  Xavier Alameda-Pineda.
\newblock Dynamical variational autoencoders: A comprehensive review.
\newblock {\em Foundations and Trends in Machine Learning}, 15(1-2):1--175,
  2021.

\bibitem{Ego4D_CVPR2022}
Kristen Grauman, Andrew Westbury, Eugene Byrne, Zachary Chavis, Antonino
  Furnari, Rohit Girdhar, Jackson Hamburger, Hao Jiang, Miao Liu, Xingyu Liu,
  et~al.
\newblock Ego4d: Around the world in 3,000 hours of egocentric video.
\newblock In {\em CVPR}, 2022.

\bibitem{ResNet_CVPR2016}
Kaiming He, Xiangyu Zhang, Shaoqing Ren, and Jian Sun.
\newblock Deep residual learning for image recognition.
\newblock In {\em CVPR}, 2016.

\bibitem{he2019bounding}
Yihui He, Chenchen Zhu, Jianren Wang, Marios Savvides, and Xiangyu Zhang.
\newblock Bounding box regression with uncertainty for accurate object
  detection.
\newblock In {\em CVPR}, 2019.

\bibitem{VPT_ECCV2022}
Menglin Jia, Luming Tang, Bor-Chun Chen, Claire Cardie, Serge Belongie, Bharath
  Hariharan, and Ser-Nam Lim.
\newblock Visual prompt tuning.
\newblock In {\em ECCV}, 2022.

\bibitem{Unct_NIPS2017}
Alex Kendall and Yarin Gal.
\newblock What uncertainties do we need in bayesian deep learning for computer
  vision?
\newblock In {\em NeurIPS}, 2017.

\bibitem{kendall2018multi}
Alex Kendall, Yarin Gal, and Roberto Cipolla.
\newblock Multi-task learning using uncertainty to weigh losses for scene
  geometry and semantics.
\newblock In {\em CVPR}, 2018.

\bibitem{DKF_arXiv2015}
Rahul~G Krishnan, Uri Shalit, and David Sontag.
\newblock Deep kalman filters.
\newblock {\em arXiv preprint arXiv:1511.05121}, 2015.

\bibitem{H2O_ICCV21}
Taein Kwon, Bugra Tekin, Jan St\"uhmer, Federica Bogo, and Marc Pollefeys.
\newblock {H2O}: Two hands manipulating objects for first person interaction
  recognition.
\newblock In {\em ICCV}, 2021.

\bibitem{RVAE_ICASSP2020}
Simon Leglaive, Xavier Alameda-Pineda, Laurent Girin, and Radu Horaud.
\newblock A recurrent variational autoencoder for speech enhancement.
\newblock In {\em ICASSP}, 2020.

\bibitem{EgoPAT3D_CVPR2022}
Yiming Li, Ziang Cao, Andrew Liang, Benjamin Liang, Luoyao Chen, Hang Zhao, and
  Chen Feng.
\newblock Egocentric prediction of action target in 3d.
\newblock In {\em CVPR}, 2022.

\bibitem{Li_ECCV2018}
Yin Li, Miao Liu, and James~M. Rehg.
\newblock In the eye of beholder: Joint learning of gaze and actions in first
  person video.
\newblock In {\em ECCV}, 2018.

\bibitem{li2021ego}
Yanghao Li, Tushar Nagarajan, Bo Xiong, and Kristen Grauman.
\newblock Ego-exo: Transferring visual representations from third-person to
  first-person videos.
\newblock In {\em CVPR}, 2021.

\bibitem{li2019learning}
Zhengqi Li, Tali Dekel, Forrester Cole, Richard Tucker, Noah Snavely, Ce Liu,
  and William~T Freeman.
\newblock Learning the depths of moving people by watching frozen people.
\newblock In {\em CVPR}, 2019.

\bibitem{liu2021egocentric}
Miao Liu, Lingni Ma, Kiran Somasundaram, Yin Li, Kristen Grauman, James~M Rehg,
  and Chao Li.
\newblock Egocentric activity recognition and localization on a 3d map.
\newblock In {\em ECCV}, 2022.

\bibitem{liu2020forecasting}
Miao Liu, Siyu Tang, Yin Li, and James~M Rehg.
\newblock Forecasting human-object interaction: joint prediction of motor
  attention and actions in first person video.
\newblock In {\em ECCV}, 2020.

\bibitem{FHOI_ECCV2020}
Miao Liu, Siyu Tang, Yin Li, and James~M Rehg.
\newblock Forecasting human-object interaction: joint prediction of motor
  attention and actions in first person video.
\newblock In {\em ECCV}, 2020.

\bibitem{OCT_CVPR2022}
Shaowei Liu, Subarna Tripathi, Somdeb Majumdar, and Xiaolong Wang.
\newblock Joint hand motion and interaction hotspots prediction from egocentric
  videos.
\newblock In {\em CVPR}, 2022.

\bibitem{malla2022nemo}
Srikanth Malla, Isht Dwivedi, Behzad Dariush, and Chiho Choi.
\newblock Nemo: Future object localization using noisy ego priors.
\newblock In {\em ITSC}, 2022.

\bibitem{nagarajan2020ego}
Tushar Nagarajan, Yanghao Li, Christoph Feichtenhofer, and Kristen Grauman.
\newblock Ego-topo: Environment affordances from egocentric video.
\newblock In {\em CVPR}, 2020.

\bibitem{nunez2022egocentric}
Adri{\'a}n N{\'u}{\~n}ez-Marcos, Gorka Azkune, and Ignacio Arganda-Carreras.
\newblock Egocentric vision-based action recognition: a survey.
\newblock {\em Neurocomputing}, 472:175--197, 2022.

\bibitem{panteleris2018using}
Paschalis Panteleris, Iason Oikonomidis, and Antonis Argyros.
\newblock Using a single rgb frame for real time 3d hand pose estimation in the
  wild.
\newblock In {\em WACV}, 2018.

\bibitem{qiu2022egocentric}
Jianing Qiu, Lipeng Chen, Xiao Gu, Frank P-W Lo, Ya-Yen Tsai, Jiankai Sun,
  Jiaqi Liu, and Benny Lo.
\newblock Egocentric human trajectory forecasting with a wearable camera and
  multi-modal fusion.
\newblock {\em IEEE Robotics and Automation Letters}, 7(4):8799--8806, 2022.

\bibitem{rangapuram2018deep}
Syama~Sundar Rangapuram, Matthias~W Seeger, Jan Gasthaus, Lorenzo Stella,
  Yuyang Wang, and Tim Januschowski.
\newblock Deep state space models for time series forecasting.
\newblock In {\em NeurIPS}, 2018.

\bibitem{ProTran_NIPS2021}
Binh Tang and David~S Matteson.
\newblock Probabilistic transformer for time series analysis.
\newblock In {\em NeurIPS}, 2021.

\bibitem{RAFT_ECCV2020}
Zachary Teed and Jia Deng.
\newblock Raft: Recurrent all-pairs field transforms for optical flow.
\newblock In {\em ECCV}, 2020.

\bibitem{Tekin_2019_CVPR}
Bugra Tekin, Federica Bogo, and Marc Pollefeys.
\newblock H+o: Unified egocentric recognition of 3d hand-object poses and
  interactions.
\newblock In {\em CVPR}, 2019.

\bibitem{thiede2019analyzing}
Luca~Anthony Thiede and Pratik~Prabhanjan Brahma.
\newblock Analyzing the variety loss in the context of probabilistic trajectory
  prediction.
\newblock In {\em CVPR}, 2019.

\bibitem{Transformer_NIPS2017}
Ashish Vaswani, Noam Shazeer, Niki Parmar, Jakob Uszkoreit, Llion Jones,
  Aidan~N Gomez, {\L}ukasz Kaiser, and Illia Polosukhin.
\newblock Attention is all you need.
\newblock In {\em NeurIPS}, 2017.

\bibitem{wang2021multi}
Jiashun Wang, Huazhe Xu, Medhini Narasimhan, and Xiaolong Wang.
\newblock Multi-person 3d motion prediction with multi-range transformers.
\newblock In {\em NeurIPS}, 2021.

\bibitem{wang2020symbiotic}
Xiaohan Wang, Yu Wu, Linchao Zhu, and Yi Yang.
\newblock Symbiotic attention with privileged information for egocentric action
  recognition.
\newblock In {\em AAAI}, 2020.

\bibitem{wiest2012probabilistic}
J{\"u}rgen Wiest, Matthias H{\"o}ffken, Ulrich Kre{\ss}el, and Klaus Dietmayer.
\newblock Probabilistic trajectory prediction with gaussian mixture models.
\newblock In {\em IVS}, 2012.

\bibitem{DSAE_ICML2018}
Li Yingzhen and Stephan Mandt.
\newblock Disentangled sequential autoencoder.
\newblock In {\em ICML}, 2018.

\bibitem{NewCRFs_CVPR2022}
Weihao Yuan, Xiaodong Gu, Zuozhuo Dai, Siyu Zhu, and Ping Tan.
\newblock {NeW CRFs}: Neural window fully-connected crfs for monocular depth
  estimation.
\newblock In {\em CVPR}, 2022.

\bibitem{yuan2019ego}
Ye Yuan and Kris Kitani.
\newblock Ego-pose estimation and forecasting as real-time {PD} control.
\newblock In {\em ICCV}, 2019.

\bibitem{AGF_ICCV21}
Ye Yuan, Xinshuo Weng, Yanglan Ou, and Kris~M Kitani.
\newblock Agentformer: Agent-aware transformers for socio-temporal multi-agent
  forecasting.
\newblock In {\em ICCV}, 2021.

\bibitem{zhang2019sr}
Pu Zhang, Wanli Ouyang, Pengfei Zhang, Jianru Xue, and Nanning Zheng.
\newblock Sr-lstm: State refinement for lstm towards pedestrian trajectory
  prediction.
\newblock In {\em CVPR}, 2019.

\bibitem{zhang2021consistent}
Zhoutong Zhang, Forrester Cole, Richard Tucker, William~T Freeman, and Tali
  Dekel.
\newblock Consistent depth of moving objects in video.
\newblock {\em ACM TOG}, 40(4):1--12, 2021.

\bibitem{Open3D_2018}
Qian-Yi Zhou, Jaesik Park, and Vladlen Koltun.
\newblock {Open3D}: {A} modern library for {3D} data processing.
\newblock {\em arXiv:1801.09847}, 2018.

\end{thebibliography}
}

\clearpage
\begin{appendix}

\section*{Appendix}

\begin{figure*}[t]
    \centering
    \includegraphics[width=\framewidth]{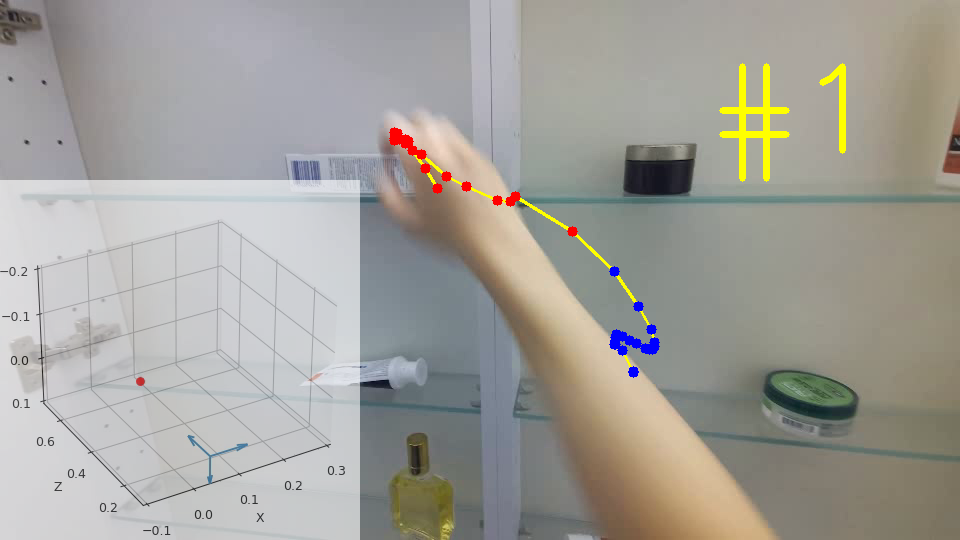}
    \includegraphics[width=\framewidth]{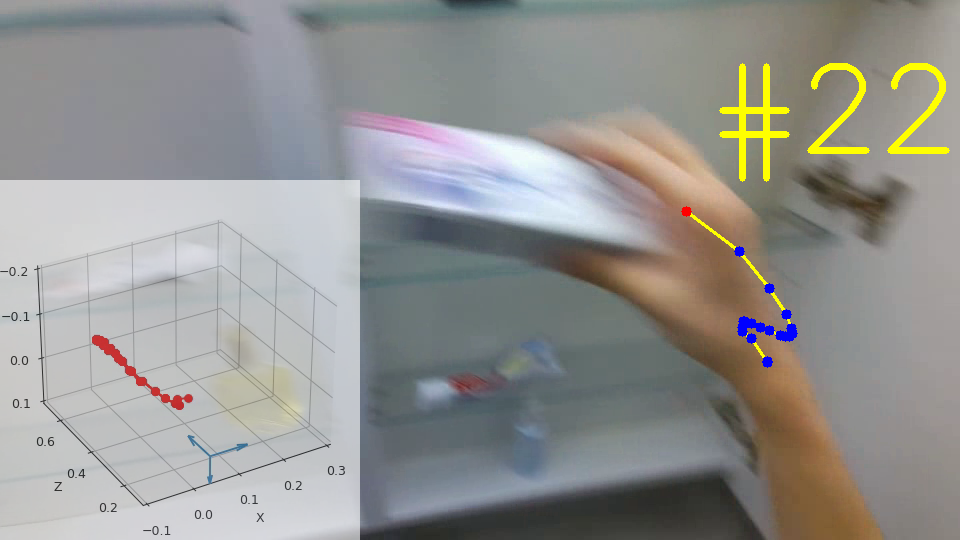}
    \includegraphics[width=\framewidth]{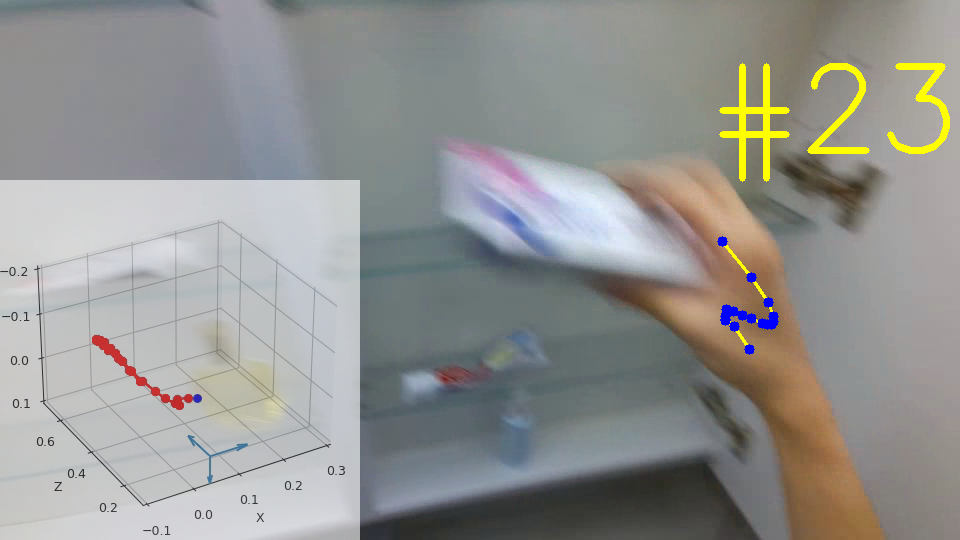}
    \includegraphics[width=\framewidth]{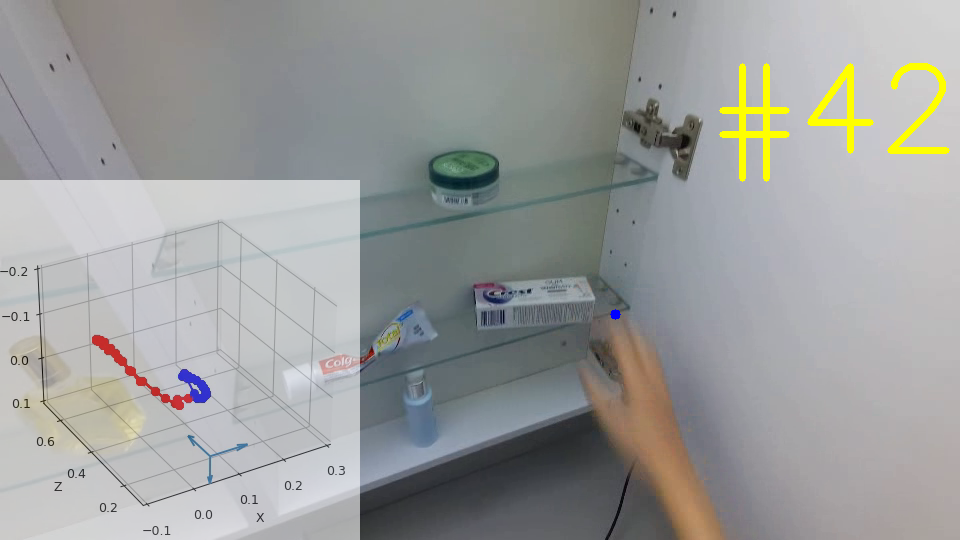}
    \includegraphics[width=\framewidth]{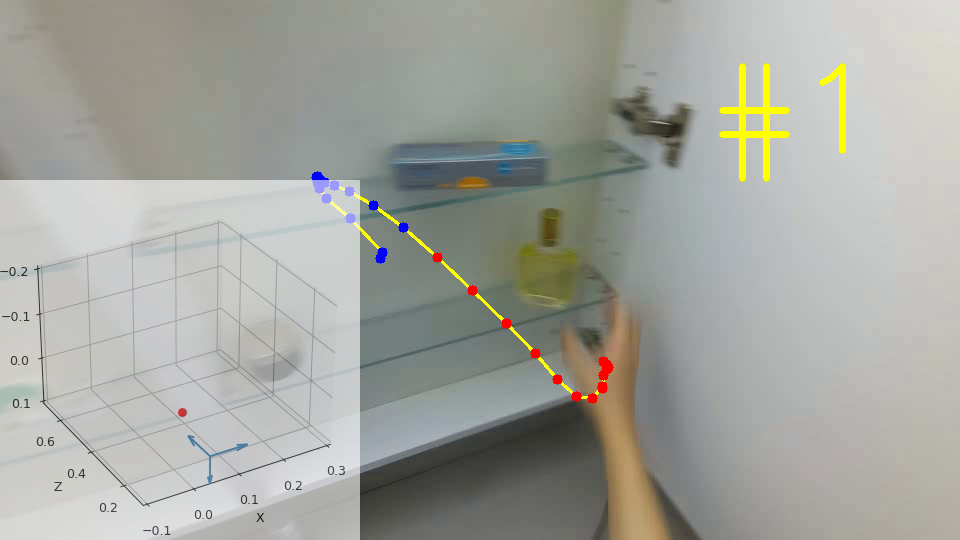}
    \includegraphics[width=\framewidth]{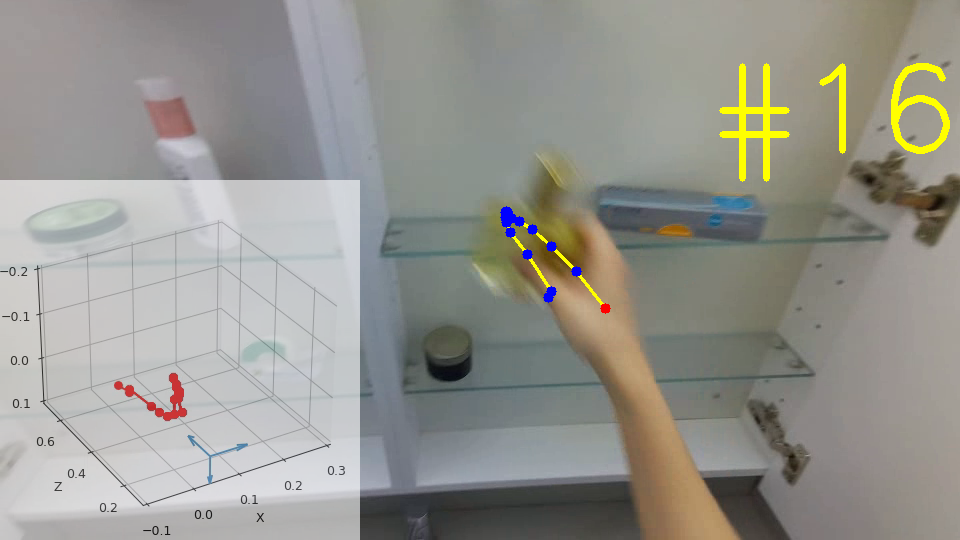}
    \includegraphics[width=\framewidth]{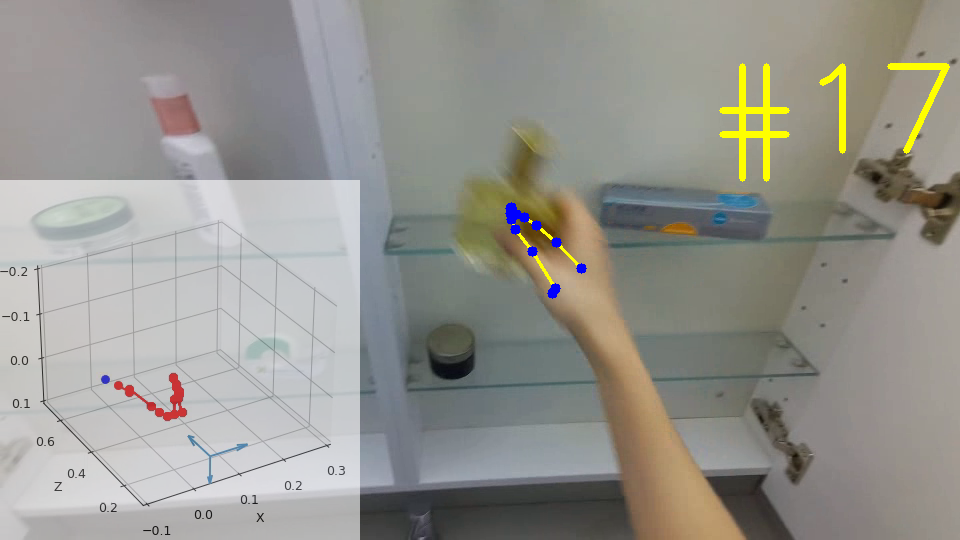}
    \includegraphics[width=\framewidth]{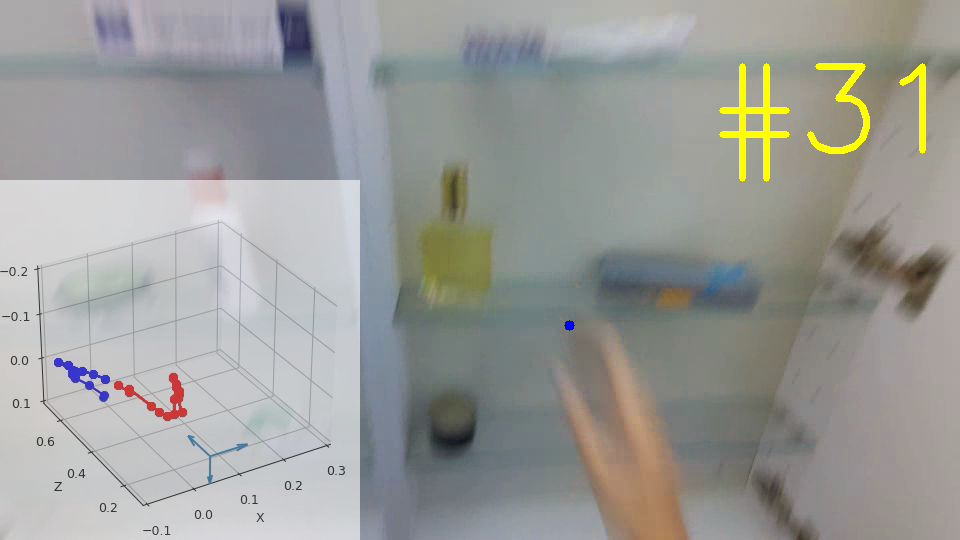}
    \includegraphics[width=\framewidth]{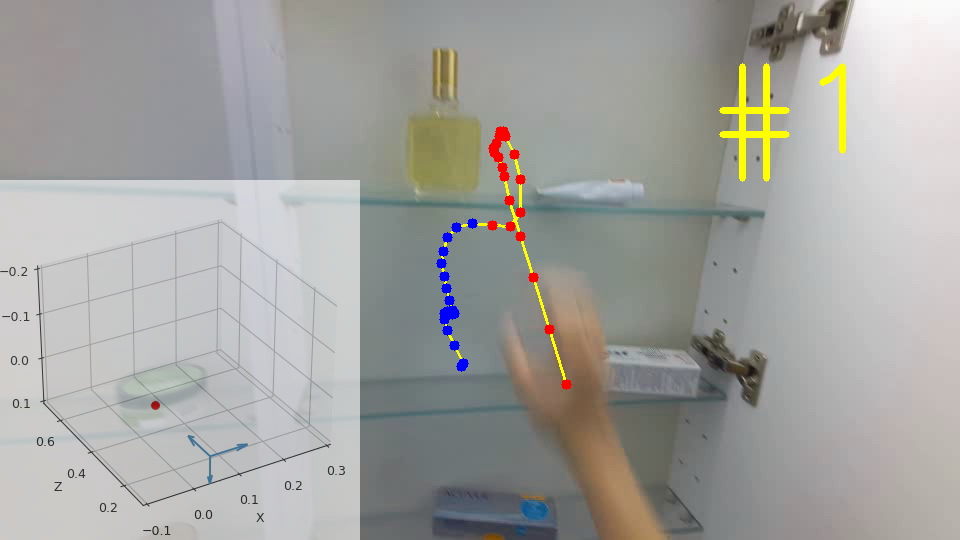}
    \includegraphics[width=\framewidth]{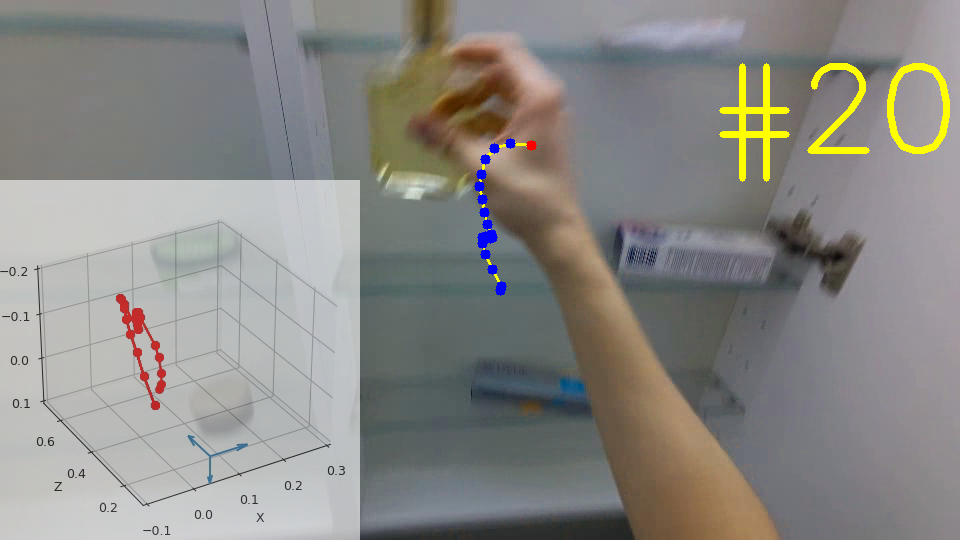}
    \includegraphics[width=\framewidth]{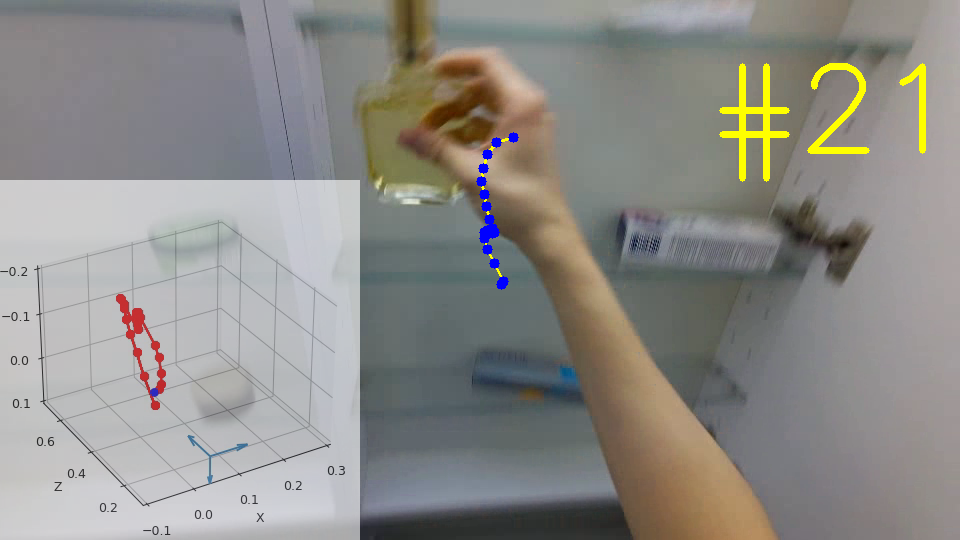}
    \includegraphics[width=\framewidth]{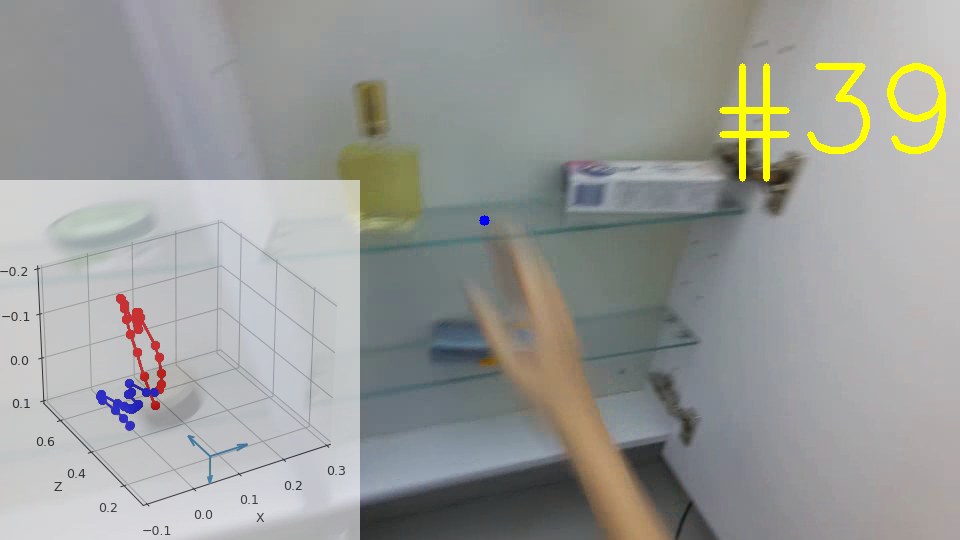}
\caption{\textbf{Dataset Examples}. For each video (in a row), the global 3D trajectory and the projected 2D trajectory are visualized, where the past and future trajectory segments are in red and blue, respectively. Zoom in for more details.}
\label{fig:data}
\end{figure*}

In this document, we provide more details of the data collection and annotation, model implementation, evaluation results, and visualizations.

\section{Details of the Datasets}

\subsection{Annotation Workflow}

Following the similar pipeline in the EgoPAT3D~\cite{EgoPAT3D_CVPR2022}, we propose to obtain the 3D hand trajectory annotations based on egocentric RGB-D recordings. In the following paragraphs, we elaborate on each processing step based on RGB-D data from the EgoPAT3D~\cite{EgoPAT3D_CVPR2022} and H2O~\cite{H2O_ICCV21}.

\paragraph{Clip Division} The EgoPAT3D dataset consists of RGB-D data of hand-object manipulation in 14 indoor scenes. We leverage the provided manual clip divisions and the hand landmarks to obtain more accurate trajectory divisions. Specifically, let $(s_m, e_m)$ denote the start and end of a manually annotated trajectory, and $\{t_s,\ldots,t_e\}$ denote the indices of detected 3D hand landmarks, our trajectory start and end are determined by $\max(s_m, t_s)$ and $\min(e_m, t_e)$, respectively. This technique could mitigate the ambiguity of trajectory start and end. Then, we use them to obtain the RGB video clips from the raw recordings. The H2O dataset contains 184 long videos and each video is annotated with 3D poses of the left and right hand as well as the binary validity flag. The trajectory start and end are determined by the validity flag.

\vspace{-2mm}
\paragraph{2D Trajectory} For each clip, we found the hand trajectory is not stable if only using the centers of frame-wise hand landmarks as trajectory points. Therefore, for the EgoPAT3D dataset, we propose to leverage the optical flow model RAFT~\cite{RAFT_ECCV2020} to warp the hand landmark center as the 2D hand trajectory. Specifically, we apply the RAFT to the forward pass starting from the first 2D location $\mathbf{p}_1$ of the hand and backward pass starting from the last location $\mathbf{p}_T$ of the hand, resulting in the forward trajectory $\{\mathbf{p}_t^{(f)}\}_{t=1}^T$ and backward trajectory $\{\mathbf{p}_t^{(b)}\}_{t=1}^T$. Then, for each frame $t$, the ultimate 2D location is determined by a temporally weighted sum $\tilde{\mathbf{p}}_t=w_t\mathbf{p}_t^{(f)} + (1-w_t)\mathbf{p}_t^{(b)}$ where the weight $w_t$ is temporally decreasing from $1.0$ to a constant $c$ by $w_t=c+(1-c)/(1+\exp(t-T/2))$. In practice, we set $c$ to 0.3. The rationale of weighing is to mitigate the error accumulation from the RAFT model. It assigns more weight to the earlier locations by forward flow and more weight to the latter locations by backward flow, with the margin $c$ between the two passes.

\begin{figure*}[t]
    \centering
    \subcaptionbox{Bathroom Cabinet}{
        \includegraphics[width=0.3\textwidth]{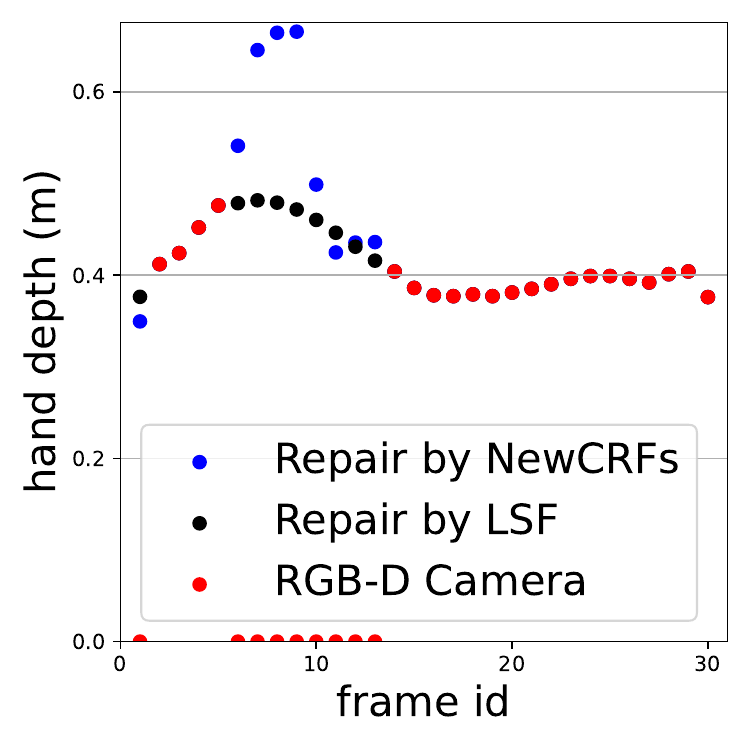}
    }
    \subcaptionbox{Bathroom Counter}{
        \includegraphics[width=0.3\textwidth]{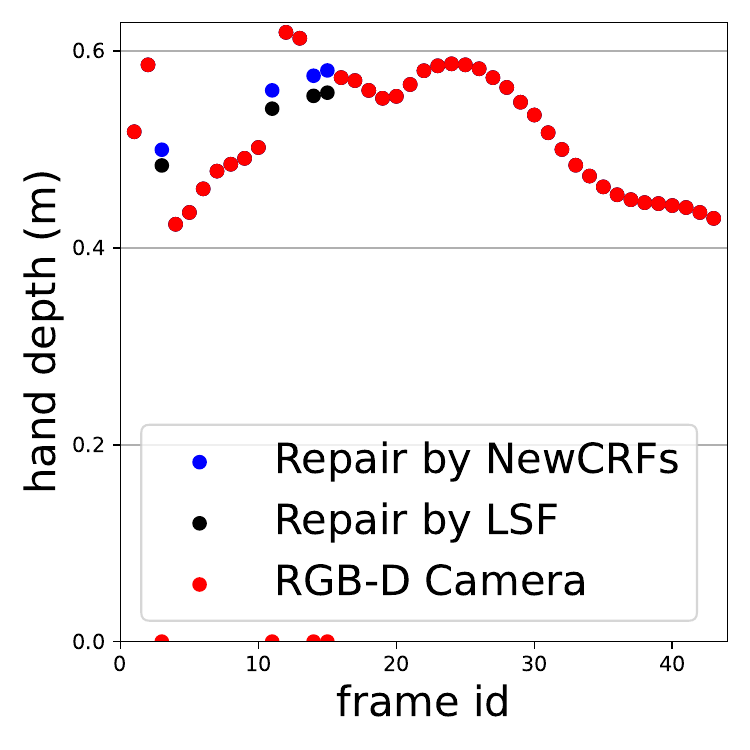}
    }
    \subcaptionbox{Bin}{
        \includegraphics[width=0.3\textwidth]{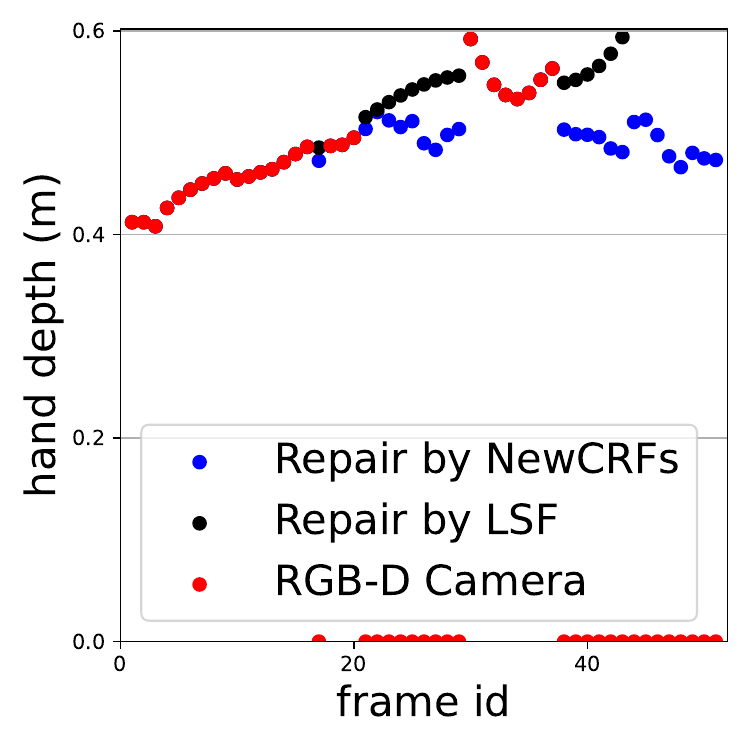}
    }
    \vfill
    \subcaptionbox{Kitchen Cupboard}{
        \includegraphics[width=0.3\textwidth]{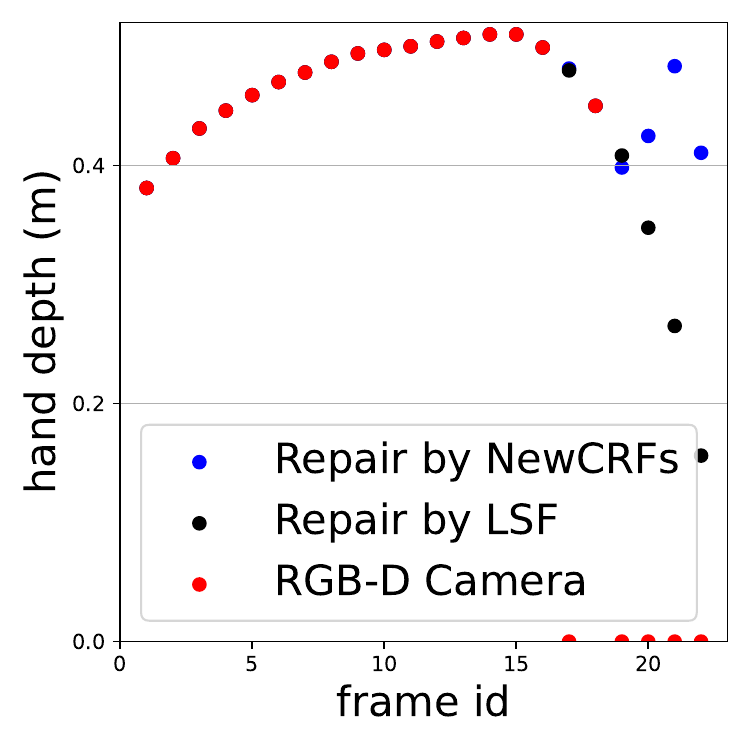}
    }
    \subcaptionbox{Microwave}{
        \includegraphics[width=0.3\textwidth]{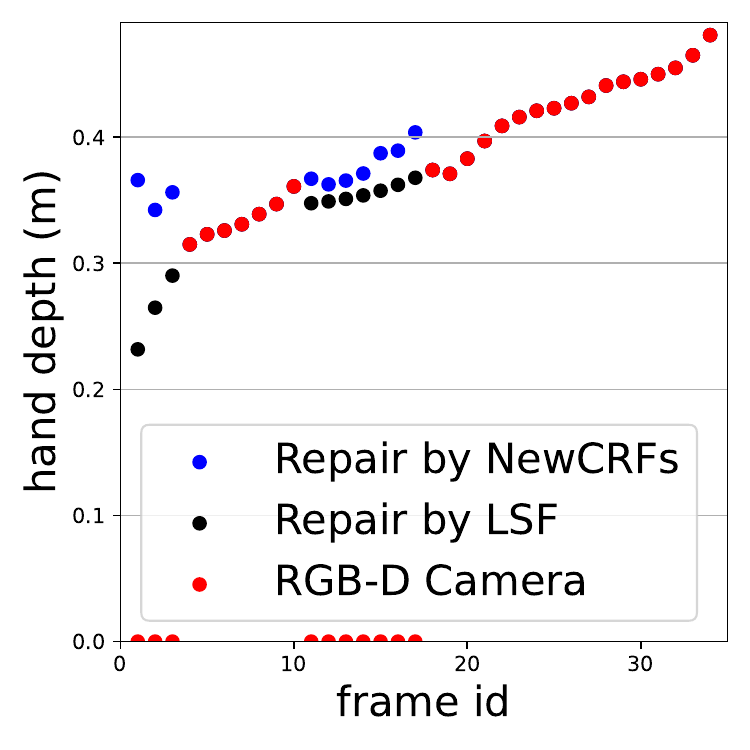}
    }
    \subcaptionbox{Nightstand}{
        \includegraphics[width=0.3\textwidth]{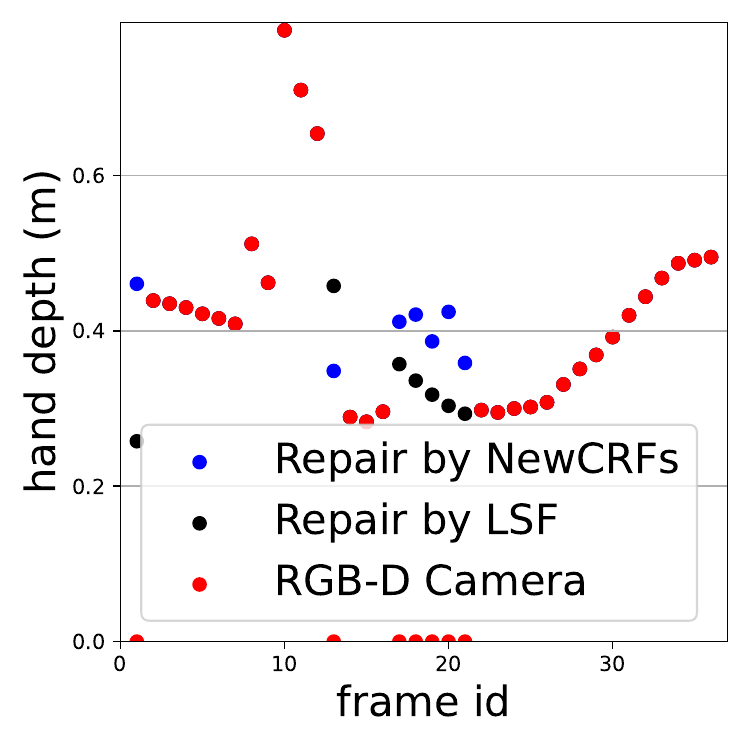}
    }
    \captionsetup{font=small,aboveskip=5pt,belowskip=0pt}
    \caption{Examples of comparison between the Least Square Fitting (LSF) and the depth estimation model NewCRFs~\cite{NewCRFs_CVPR2022} for repairing the noisy depth values from EgoPAT3D RGB-D data. It's clear that on this video dataset with dynamic background, a simple LSF with a multinomial model could achieve a much better depth repairing effect than the state-of-the-art deep learning model NewCRFs.}
    \label{supfig:depth}
\end{figure*}

\paragraph{Local 3D Trajectory} With the 2D hand trajectory, it is straightforward to obtain the 3D hand trajectory by fetching the depth of each trajectory point from the RGB-D clips. However, we noticed that due to the fast motion of the hand and camera, the recorded depth channels in those frames could be missing, i.e., depth values are zeros (see the red dots in Fig.~\ref{supfig:depth}). To obtain high-quality 3D hand trajectory annotations, we initially attempted to use the state-of-the-art depth estimation model NewCRFs~\cite{NewCRFs_CVPR2022} to estimate the missing depths from RGB frames. However, it cannot work well due to the camera motion that results in dynamic scenes in RGB frames. Instead, we found that a simple least-square fitting (LSF) by combining the third-order polynomial and sine functions, i.e., $z_{\alpha}(t)=\alpha_1t^3 + \alpha_2t^2 + \alpha_3t + \alpha_4 + \alpha_5\sin(\alpha_6 t)$, could repair the missing depth. For both EgoPAT3D and H2O, we apply the LSF to repair 3D hand trajectory depth. To enable successful depth fitting, we use at least 10 valid trajectory points to fit a multinomial model on each 3D hand trajectory that contains invalid depths.

\vspace{-2mm}
\paragraph{Global 3D Trajectory} Note that the 3D trajectory points from the previous step are defined in the local camera coordinate system. When the camera is moving in an egocentric view, using RGB videos to predict the local 3D trajectory will be ambiguous. In other words, distinct visual contents are forced to learn to predict numerically similar coordinates. To eliminate the ambiguity, similar to EgoPAT3D~\cite{EgoPAT3D_CVPR2022}, we propose to transform the 3D trajectory targets into a global world coordinate system with reference to the first frame. This is a visual odometry procedure that computes the 3D homogeneous transformation $\mathbf{M}_t\in\mathbb{R}^{4\times 4}$ between camera poses at two successive frames $t-1$ and $t$. Eventually, a local 3D trajectory point $\mathbf{p}_t^{l}$ is transformed as a global 3D trajectory point $\mathbf{p}_t^{g}$ by the accumulative matrix product $\mathbf{p}_t^{g}=\prod_{k=1}^t\mathbf{M}_k\mathbf{p}_t^{l}$. In experiments, we use the global 3D trajectory $\{\mathbf{p}_t^{g}\}_{t=1}^T$ as the ground truth for model training, evaluation, and visualization by default. Fig.~\ref{fig:data} shows three video examples with global 3D trajectory annotations.

\begin{table}[htpb]
    \centering
    \small
    \setlength{\tabcolsep}{1.2mm}
    \setlength{\extrarowheight}{0.5mm}
    \captionsetup{font=small,aboveskip=3pt,belowskip=-5pt}
    \caption{Summary of camera intrinsics}
    \begin{tabular}{l|l|c}
        \toprule
         \multirow{ 3}{*}{EgoPAT3D} & resolution  &  $H=2160$, \;\; $W=3840$\\
         & focal length & $f_x=1808.203$, \;\; $f_y=1807.946$ \\
         & principle point & $o_x=1942.287$, \;\; $o_y=1123.822$ \\
         \hline
         \multirow{ 3}{*}{H2O} & resolution  &  $H=720$, \;\; $W=1280$\\
         & focal length & $f_x=636.659$, \;\; $f_y=636.252$ \\
         & principle point & $o_x=635.284$, \;\; $o_y=366.874$ \\
         \bottomrule
    \end{tabular}
    \label{suptab:intrinsics}
\end{table}

\subsection{Camera Intrinsics and Poses}

For both the EgoPAT3D and H2O, the camera intrinsics are fixed across all samples. Table~\ref{suptab:intrinsics} summarizes the camera intrinsics of the dataset we used in this paper. Note that the intrinsics are scaled with the factor 0.25 when we down-scale the RGB videos to the input resolution. For camera poses of EgoPAT3D, we use Open3D~\cite{Open3D_2018} library to perform visual odometry\footnote{In practice, we followed the EgoPAT3D to use the Open3D API (\emph{RGBDOdometryJacobianFromHybridTerm}) to compute the 3D camera motion.} by using adjacent RGB-D pairs so that the camera motion is obtained. The camera poses of H2O dataset are given for each video frame.

\section{Additional Implementation Details}

\paragraph{Data Structure} To enable efficient parallel training with batches of data input that contain videos of varying lengths, we adopt the mask mechanism in our implementation. Specifically, we set the maximum length of each video to 40 and 64 for EgoPAT3D and H2O, respectively. The lengths of the past observation and future frames are determined by the actual video length. For instance, when the observation ratio is set to 0.6, a sample with 35 frames in total has 21 observed frames, 14 unobserved frames, and 5 zero-padded frames. Since the visual background of RGB videos is relatively clean, we resize videos into the size of $64 \times 64$ in training and inference.

\paragraph{Model Structure} For the ResNet-18 backbone, we replace the global pooling layer after the last residual block with $\texttt{torch.flatten}$, in order to preserve as much visual contextual information as possible. When the visual prompt tuning (VPT) is utilized, the width of the padded learnable pixels is set to 5 as suggested by~\cite{VPT_ECCV2022}, resulting in 1380\footnote{For $64\times 64$ input, the number of learnable parameters in prompt embeddings is computed by $(64+5\times2)^2 - 64^2=1380$.} additional parameters to learn. For the ViT backbone, we adopt the $\texttt{vit/b16-224}$ architecture provided by TIMM, which is pre-trained on the ImageNet-21K dataset. For either ResNet-18 or ViT-based frame encoder $f_{\mathcal{V}}$, the output feature is embedded by a two-layer MLP with 512 and 256 hidden units. For the trajectory encoder $f_{\mathcal{T}}$, we use a two-layer MLP with 128 and 256 hidden units. For both visual and trajectory transformer encoders, we utilize the standard transformer encoder architecture, which consists of 6 multi-head self-attention blocks where the number of heads is 8 and the MLP ratio is 4. For the decoder, we implement the three prediction branches, i.e., future trajectory prediction, uncertainty prediction, and velocity prediction, using three MLP heads, each of which consists of 128 and 3 hidden units. For trajectory and velocity prediction outputs, we use $\texttt{tanh}$ activation, while for the uncertainty output, we use $\texttt{softplus}$ activation. Besides, for the velocity prediction, layer normalization is applied to each hidden layer.

\begin{table*}[t]
\centering
\captionsetup{font=small,aboveskip=3pt}
\caption{\textbf{Results of models training on H2O-DT dataset}. We report all results of models trained by annotations from H2O-DT (left) and its version without depth repair (right), and \underline{tested on the accurate H2O-PT test set}. All models are built with ResNet-18 backbone. Best and secondary results are viewed in bold \textbf{black} and \textcolor{blue}{\textbf{blue}} colors, respectively.
}
\label{tab:h2o_dt}
\small
\setlength{\tabcolsep}{1.7mm}
\setlength{\extrarowheight}{0.5mm}
    \begin{tabular}{l|ccc|ccc?ccc|ccc}
    \toprule
    \multicolumn{1}{c|}{\multirow{3}{*}{Models}} & \multicolumn{6}{c?}{H2O-DT} & \multicolumn{6}{c}{H2O-DT (w/o depth repair)} \\
        \cline{2-13}  & \multicolumn{3}{c|}{ADE ($\downarrow$)} 
        & \multicolumn{3}{c?}{FDE ($\downarrow$)} & \multicolumn{3}{c|}{ADE ($\downarrow$)} 
        & \multicolumn{3}{c}{FDE ($\downarrow$)} \\
        \cline{2-13}          & $\text{3D}_{(\textcolor{red}{3D})}$ & $\text{2D}_{(\textcolor{red}{3D})}$  & $\text{2D}_{(\textcolor{blue}{2D})}$  & $\text{3D}_{(\textcolor{red}{3D})}$ & $\text{2D}_{(\textcolor{red}{3D})}$  & $\text{2D}_{(\textcolor{blue}{2D})}$  & $\text{3D}_{(\textcolor{red}{3D})}$ & $\text{2D}_{(\textcolor{red}{3D})}$  & $\text{2D}_{(\textcolor{blue}{2D})}$  & $\text{3D}_{(\textcolor{red}{3D})}$ & $\text{2D}_{(\textcolor{red}{3D})}$  & $\text{2D}_{(\textcolor{blue}{2D})}$\\
    \hline
    DKF~\cite{DKF_arXiv2015}  & 0.236 & 0.235 & 0.269	& 0.138 &	\textbf{0.030}	& \textbf{0.020}   &  0.199	& 0.186	& 0.208	& 0.181	& 0.153	& 0.187\\
    RVAE~\cite{RVAE_ICASSP2020} & 0.125	& 0.209	& 0.094	& 0.057	& 0.082	& 0.047  &   \textcolor{blue}{0.051}	& 0.060	& 0.059	& \textcolor{blue}{0.058}	& 0.071	& 0.059 \\
    DSAE~\cite{DSAE_ICML2018} & 0.081	& 0.113	& 0.078	& 0.043	& 0.059	& 0.040  &   0.072	& 0.068	& 0.063	& 0.067	& 0.047	& 0.077 \\
    STORN~\cite{STORN_NIPSW2014} & 0.091	& 0.100	& 0.070	& 0.245	& 0.078	& 0.040  & 0.067	& 0.061	& 0.054	& 0.135	& 0.097	& 0.121\\
    VRNN~\cite{VRNN_NIPS2015} & \textcolor{blue}{0.080}	& 0.092	& \textcolor{blue}{0.068}	& \textcolor{blue}{0.042}	& 0.035	& \textcolor{blue}{0.039}    &  0.065	& 0.063	& \textcolor{blue}{0.054}	& 0.133	& 0.087	& 0.087 \\
    SRNN~\cite{SRNN_NIPS2016} & 0.087	& 0.097	& 0.076	& 0.124	& 0.072	& 0.045   & 0.055	& \textcolor{blue}{0.059}	& 0.061	& 0.083	& 0.089	& 0.135 \\
    \hline
    AGF~\cite{AGF_ICCV21} & 0.108	& \textcolor{blue}{0.065}	& 0.080	& 0.171	& 0.061	& 0.214   & 0.099  &  0.075 & 0.065  & 0.186  & \textcolor{blue}{0.044}  & 0.056 \\ 
    OCT~\cite{OCT_CVPR2022} & 0.360	& 0.473	& 0.350	& 0.348	& 0.362	& 0.520 &  0.381	& 0.519  & 0.403  &  0.403	& 0.521  & 0.505 \\
    ProTran~\cite{ProTran_NIPS2021} & \textcolor{blue}{0.080}	& 0.082	& 0.099	& \textbf{0.023}	& \textcolor{blue}{0.031}	& 0.107  & 0.070	& 0.093  & 0.064  & 0.162	& 0.146 & \textcolor{blue}{0.041} \\
    \hline
    USST & \textbf{0.033}	& \textbf{0.041}	& \textbf{0.041}	& 0.052	& 0.050	& 0.041  & \textbf{0.032}	& \textbf{0.041}	& \textbf{0.040}	& \textbf{0.053}	& \textbf{0.041}	& \textbf{0.041}  \\
    \bottomrule
    \end{tabular}%
\end{table*}%

\paragraph{Learning and Inference} In training, we set the $\delta$ parameter of Huber loss to $1e-5$, and set the $\gamma$ coefficient of the velocity-based warping loss to 0.1. For the cosine learning rate scheduler, we adopt warm-up training in the first 10 epochs. For 500 training epochs in total, our model training can be completed within 5 hours on a single RTX A6000 GPU. In testing, we evaluate the predicted 3D trajectory in the global coordinate system by referring to the camera at the first time step, while visualizing the 2D trajectory by first projecting the global 3D trajectory into the local 3D trajectory, and then projecting the local 3D coordinates onto a video frame as 2D pixel coordinates.

\balance

\section{Additional Evaluation Results} 

\begin{table}[t]
\centering
\captionsetup{font=small,aboveskip=3pt}
\caption{\textbf{FDE results of 2D hand trajectory forecasting}. Compared models are built with ResNet-18 (R18) backbone. Best and secondary results are in bold \textbf{black} and \textcolor{blue}{blue} colors, respectively.}
\small
\setlength{\tabcolsep}{7mm}
    \begin{tabular}{l|cc}
    \toprule
    Model & Seen ($\downarrow$) & Unseen ($\downarrow$) \\
    \hline
    DKF~\cite{DKF_arXiv2015}   &   0.150     &  0.239 \\
    RVAE~\cite{RVAE_ICASSP2020} & 0.152      &  0.201   \\
    DSAE~\cite{DSAE_ICML2018}  &  0.144       &  0.233   \\
    STORN~\cite{STORN_NIPSW2014} &  0.145    & 0.266    \\
    VRNN~\cite{VRNN_NIPS2015} &   0.155    &  0.237 \\
    SRNN~\cite{SRNN_NIPS2016} &   0.157    &   0.198 \\
    \hline
    OCT~\cite{OCT_CVPR2022}    &  0.090      & 0.147   \\
    ProTran~\cite{ProTran_NIPS2021}  &  0.134    &  \textbf{0.049}  \\
    \hline
    USST (R18) &  \textcolor{blue}{0.075}   &  \textcolor{blue}{0.107} \\  
    USST (ViT)  &  \textbf{0.066}   &  0.114  \\
    \bottomrule
    \end{tabular}%
  \label{suptab:fde_compare}%
\end{table}%

\paragraph{Full results on H2O-DT} We additionally provide full experimental results by training models on \textbf{H2O-DT} and \textbf{H2O-DT} w/o depth repair in Table~\ref{tab:h2o_dt}. It shows that our USST method could still achieve the best performance using training data with inaccurate trajectory annotations.

\paragraph{FDE results on EgoPAT3D-DT}
We additionally provide the Final Displacement Error (FDE) results for 2D hand trajectory forecasting as shown in Table~\ref{suptab:fde_compare}. Our method could achieve the best performance on the seen test data while being competitive on the unseen test data. Besides, ProTran shows the best result on the unseen data, which could be attributed to its extra trajectory supervision from the full observation of the latent Gaussian distributions. 

\begin{figure}[t]
    \centering
    \includegraphics[width=\linewidth]{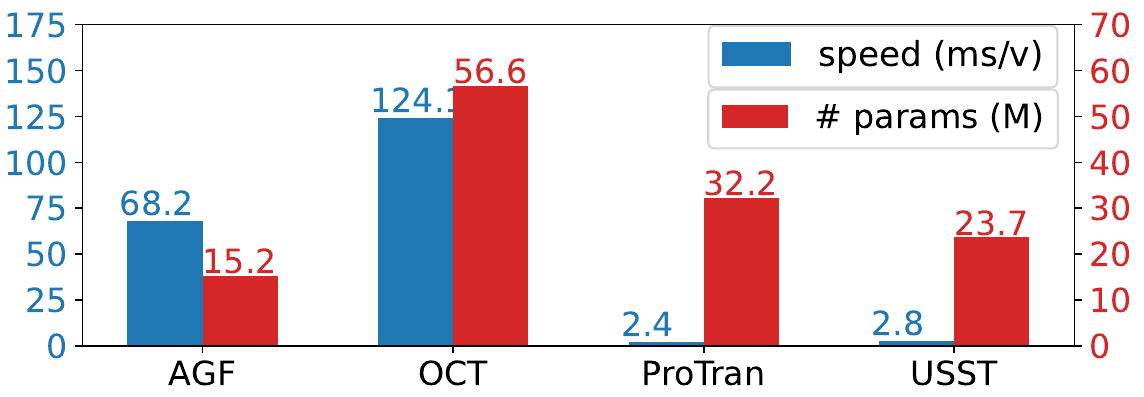}
    \captionsetup{font=small,aboveskip=0mm,belowskip=0mm}
    \vspace*{-7mm}
    \caption{Inference speed in milliseconds/video (ms/v) and the number of model parameters in million (M), tested on a single RTX 6000Ada GPU with input video size $64 \times 64 \times 64$.}
    \label{fig:speed}
\end{figure}

\paragraph{Inference speed} In Fig.~\ref{fig:speed}, we compared with the Transformer-based methods. It shows the USST achieves competitive speed to ProTran while comparable model size to AGF. With certain improvements, our method could potentially benefit the rendering latency in AR/VR.

\section{Additional Demos}

\begin{figure*}
\footnotesize
\centering
\renewcommand{\tabcolsep}{0.7pt} %
\begin{tabular}{ccc}
\parbox[t]{4mm}{\multirow{1}{*}[11em]{\rotatebox[origin=c]{90}{Small Bins (seen)}}} &
\includegraphics[width=\newfigwidth]{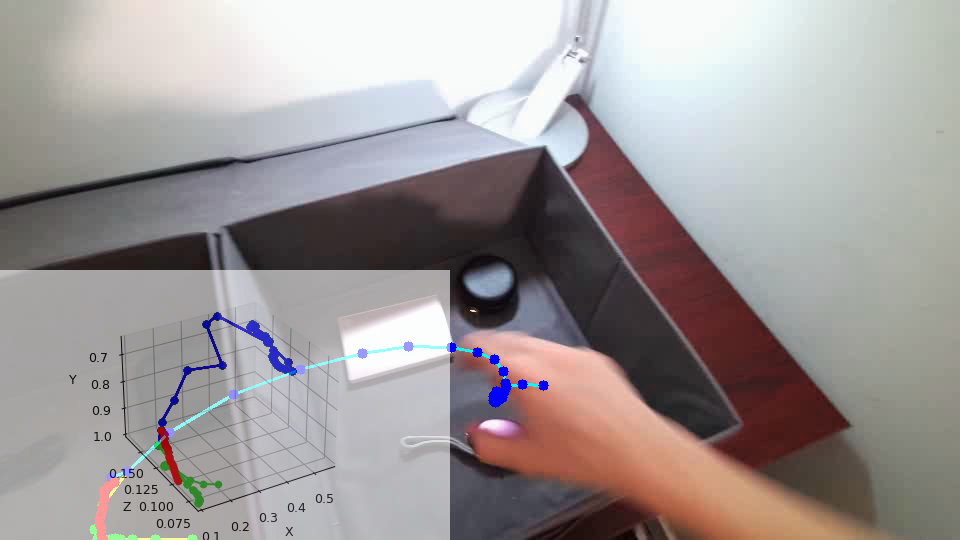} &
\includegraphics[width=\newfigwidth]{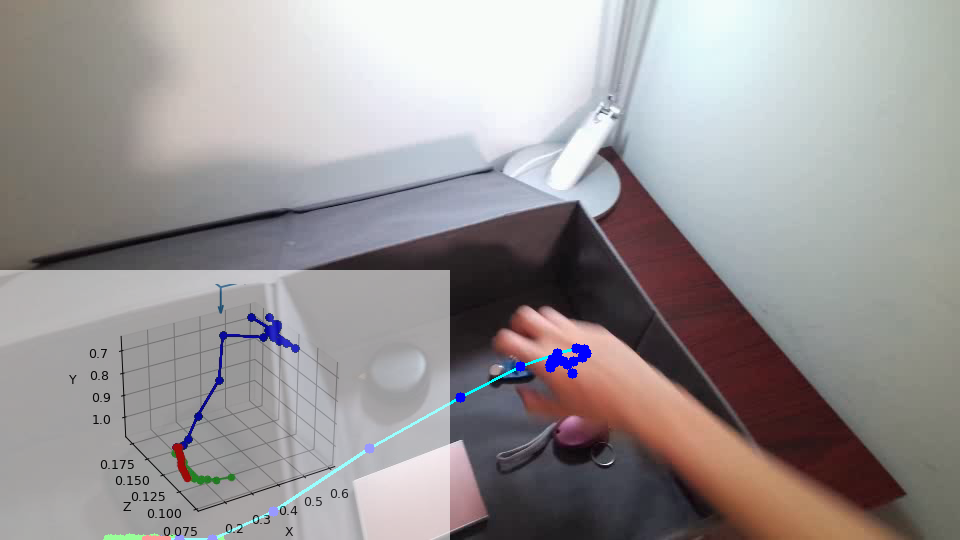} 
\\
\parbox[t]{4mm}{\multirow{1}{*}[11em]{\rotatebox[origin=c]{90}{Microwave (seen)}}} &
\includegraphics[width=\newfigwidth]{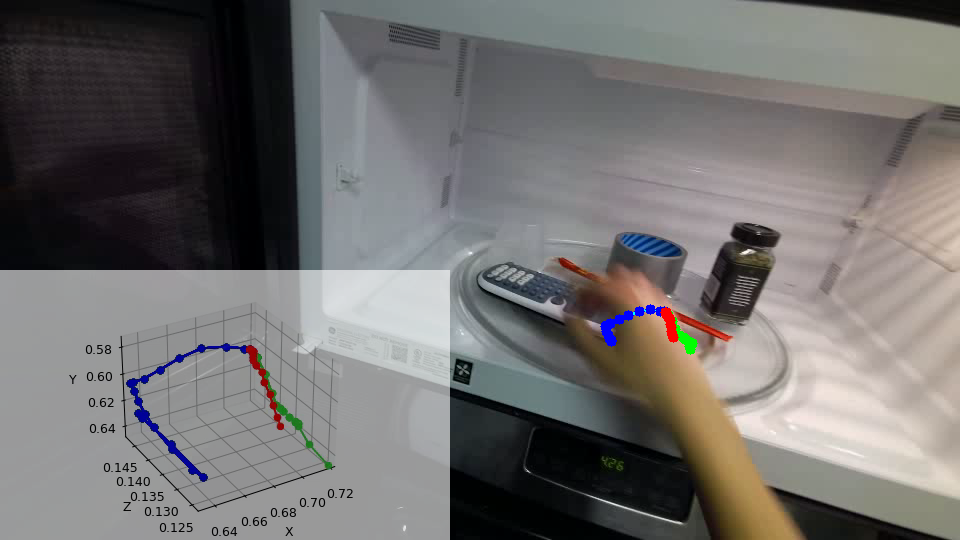} &
\includegraphics[width=\newfigwidth]{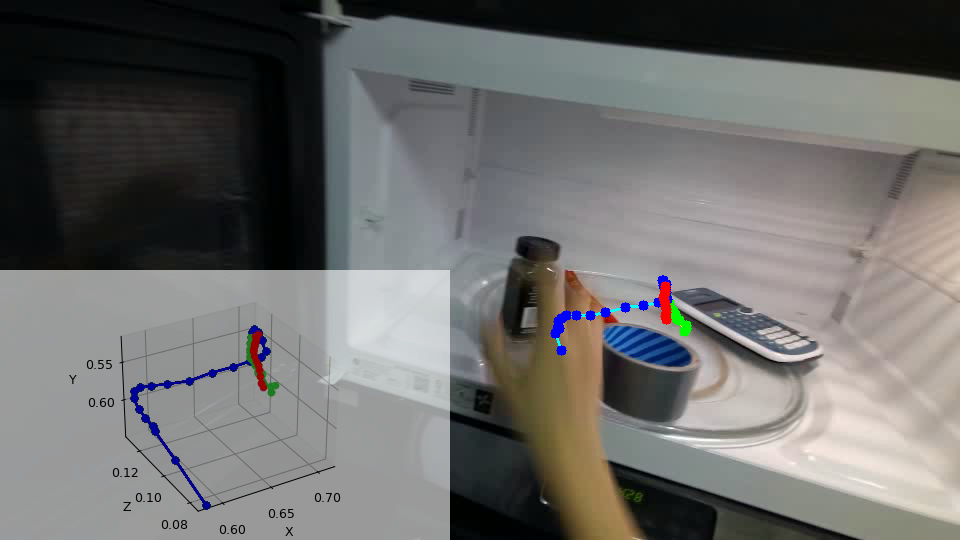} 
\\
\parbox[t]{4mm}{\multirow{1}{*}[11em]{\rotatebox[origin=c]{90}{Bathroom Cabinet (seen)}}} &
\includegraphics[width=\newfigwidth]{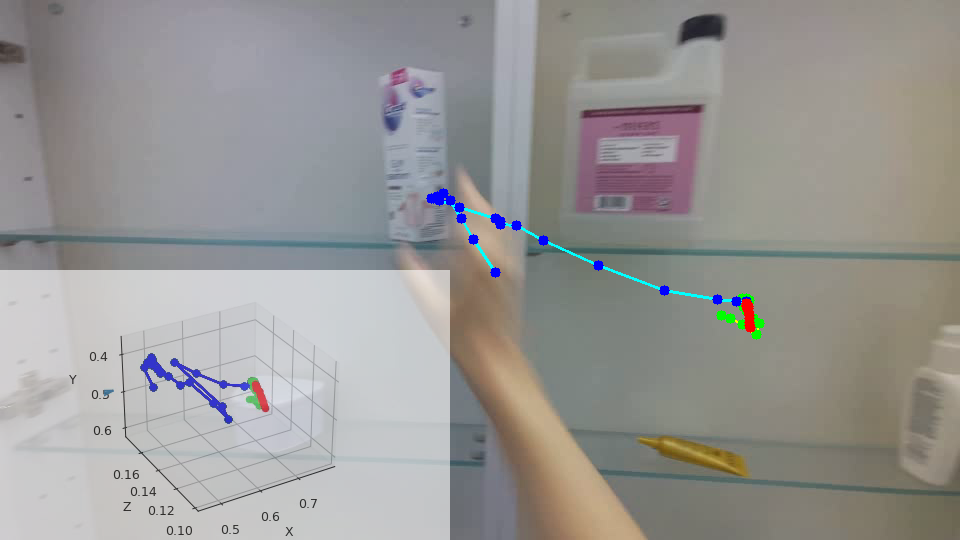} &
\includegraphics[width=\newfigwidth]{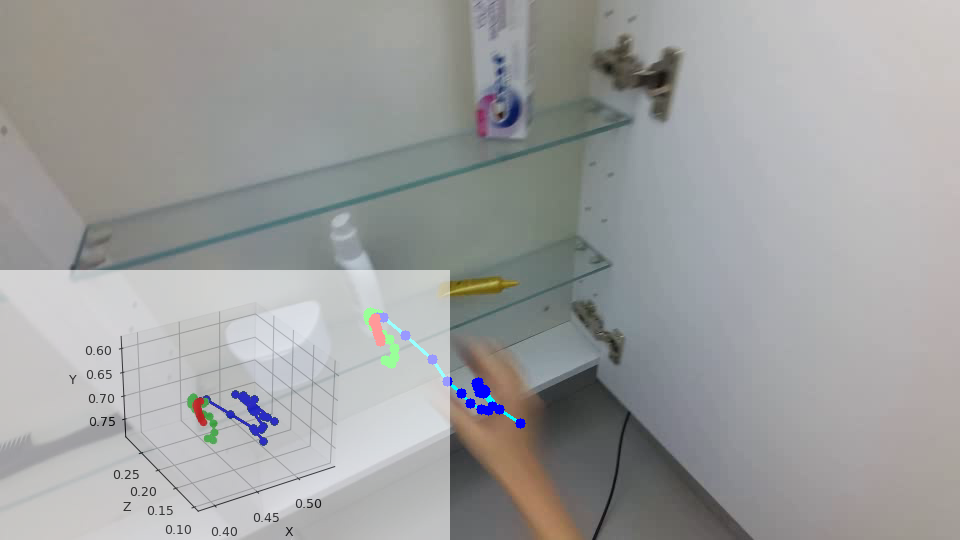} 
\\
\parbox[t]{4mm}{\multirow{1}{*}[11em]{\rotatebox[origin=c]{90}{Bathroom Counter (seen)}}} &
\includegraphics[width=\newfigwidth]{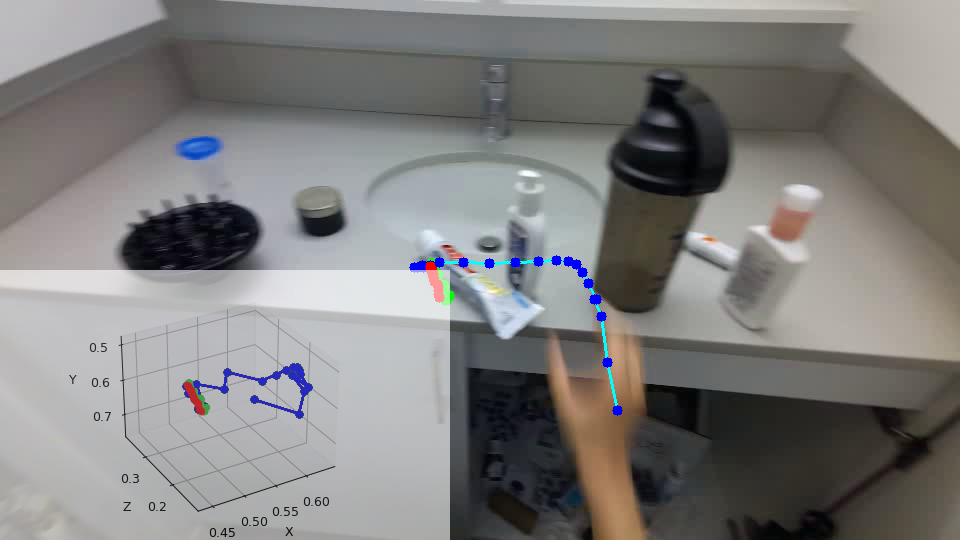} &
\includegraphics[width=\newfigwidth]{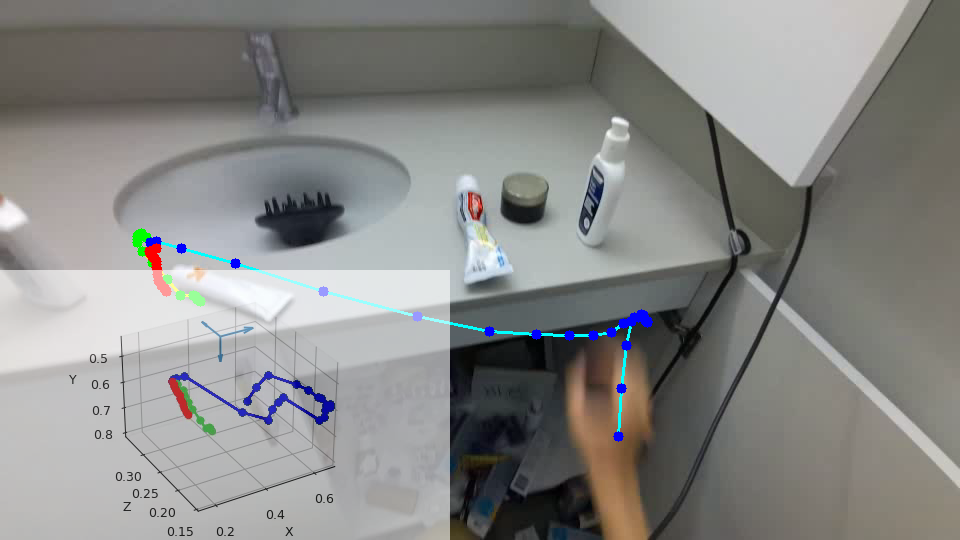} 
\\
\parbox[t]{4mm}{\multirow{1}{*}[11em]{\rotatebox[origin=c]{90}{Pantry Shelf (seen)}}} &
\includegraphics[width=\newfigwidth]{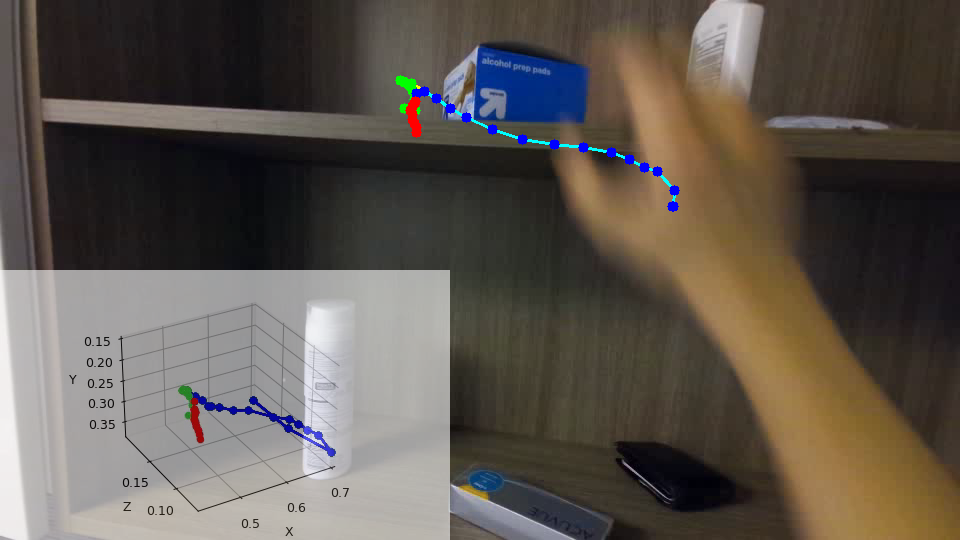} &
\includegraphics[width=\newfigwidth]{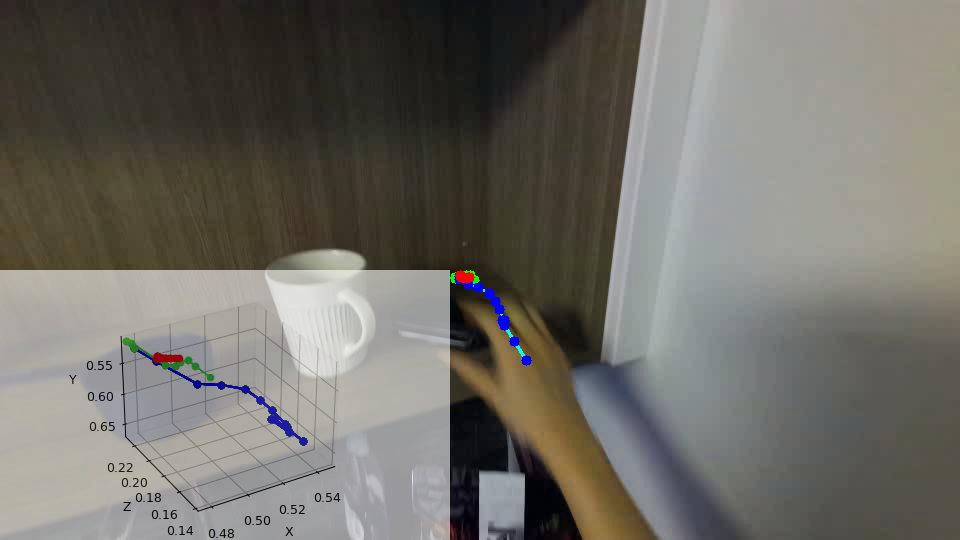} 
\\
\end{tabular}
\captionsetup{font=small,aboveskip=3pt}
\caption{\textbf{Visualization on Seen data.} For each scene (in a row), we show two examples of the 2D and 3D trajectories on the first frame. The \textcolor{blue}{blue}, \textcolor{green}{green}, and \textcolor{red}{red} trajectory points represent the past observed, future ground truth, and future predictions, respectively. 
}
\label{supfig:vis_seen}
\vspace{-10pt}
\end{figure*}

\begin{figure*}[t]
\footnotesize
\centering
\renewcommand{\tabcolsep}{0.4pt} %
\begin{tabular}{ccc}
\parbox[c]{4mm}{\multirow{1}{*}[11em]{\rotatebox[origin=c]{90}{StoveTop (unseen)}}} &
\includegraphics[width=\newfigwidth]{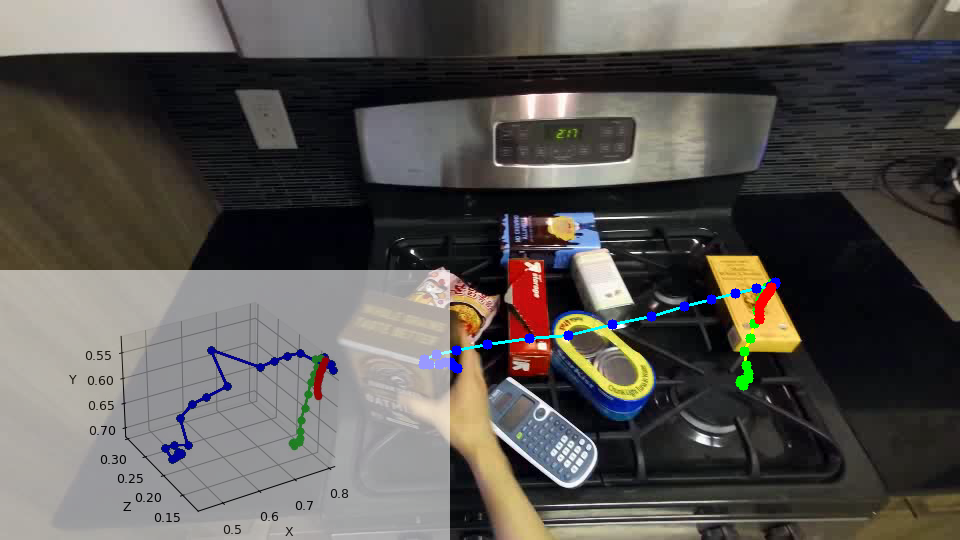} &
\includegraphics[width=\newfigwidth]{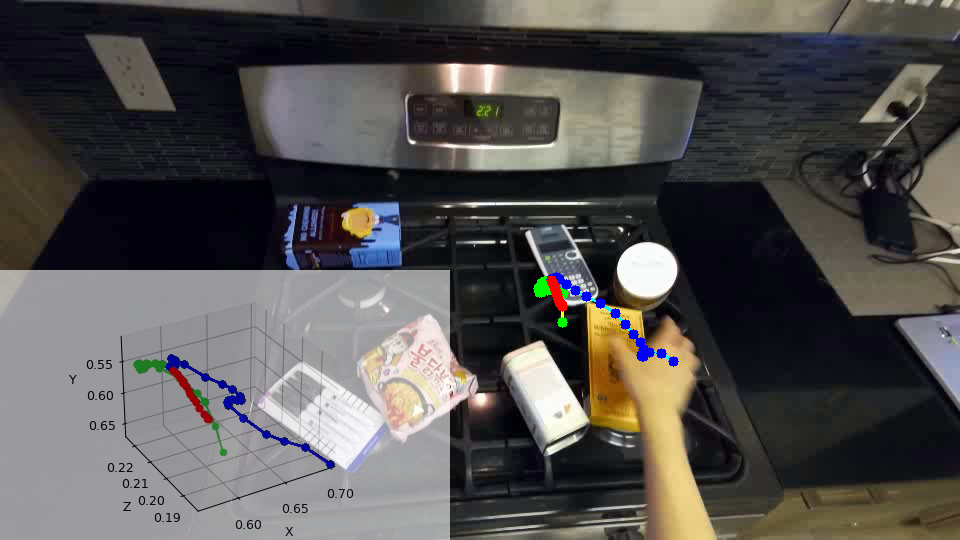}
\\
\parbox[c]{4mm}{\multirow{1}{*}[12em]{\rotatebox[origin=c]{90}{Wooden Table (unseen)}}} &
\includegraphics[width=\newfigwidth]{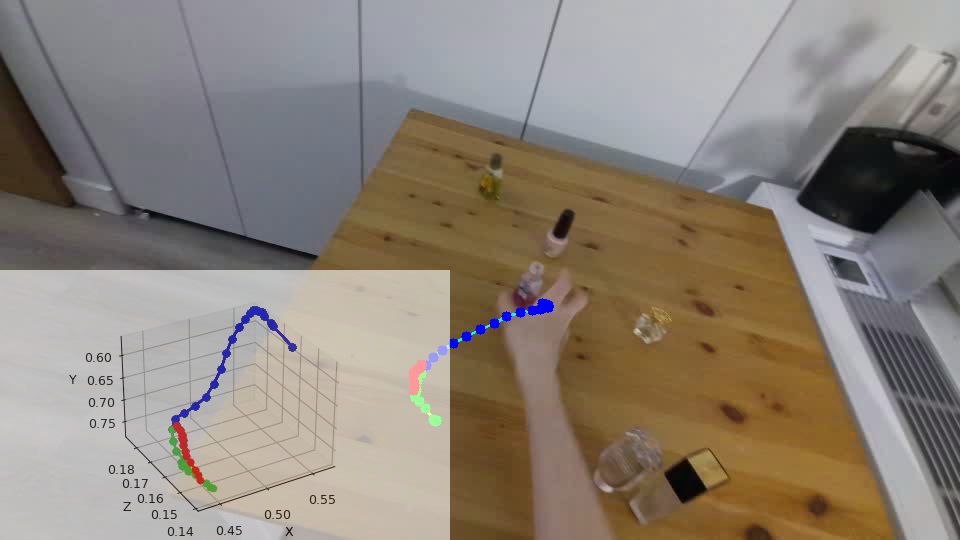} &
\includegraphics[width=\newfigwidth]{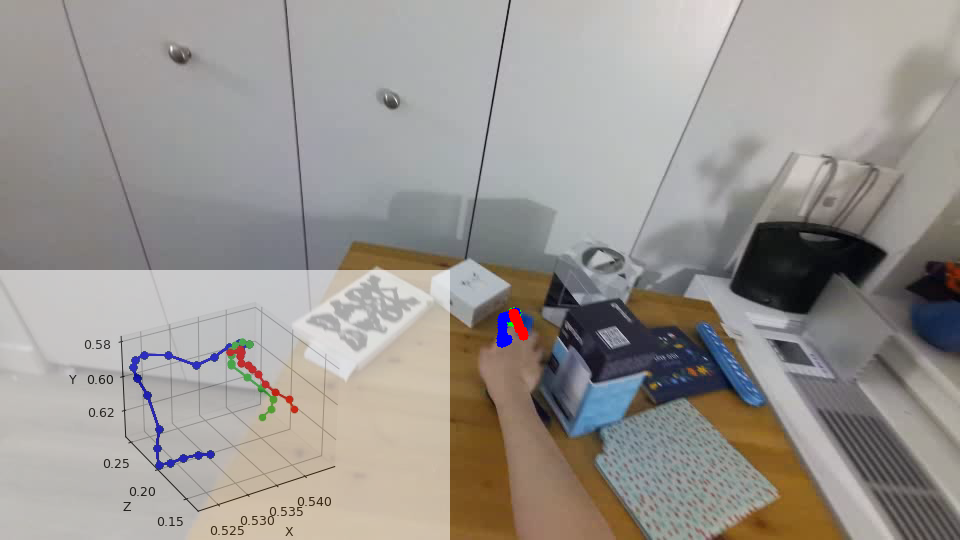} 
\\
\parbox[c]{4mm}{\multirow{1}{*}[12em]{\rotatebox[origin=c]{90}{Windowsill AC (unseen)}}} &
\includegraphics[width=\newfigwidth]{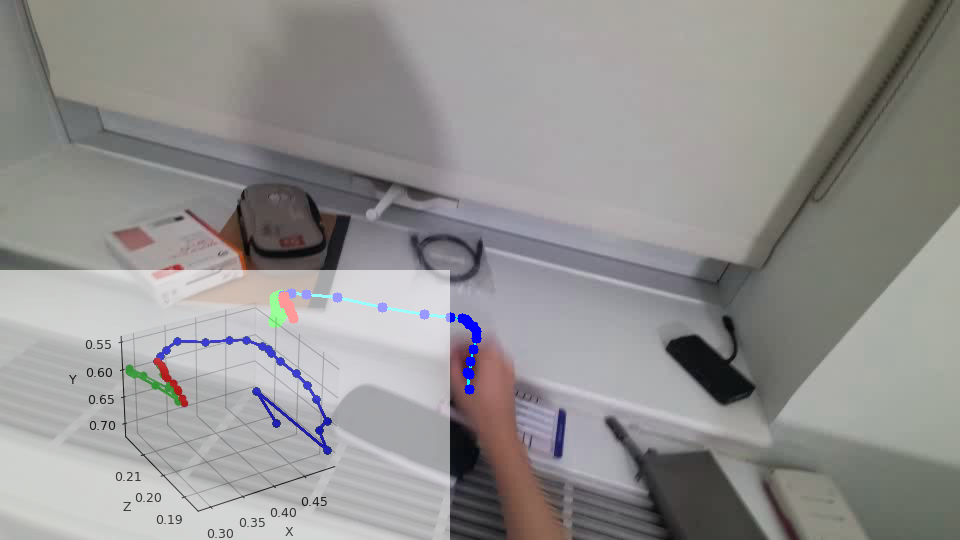} &
\includegraphics[width=\newfigwidth]{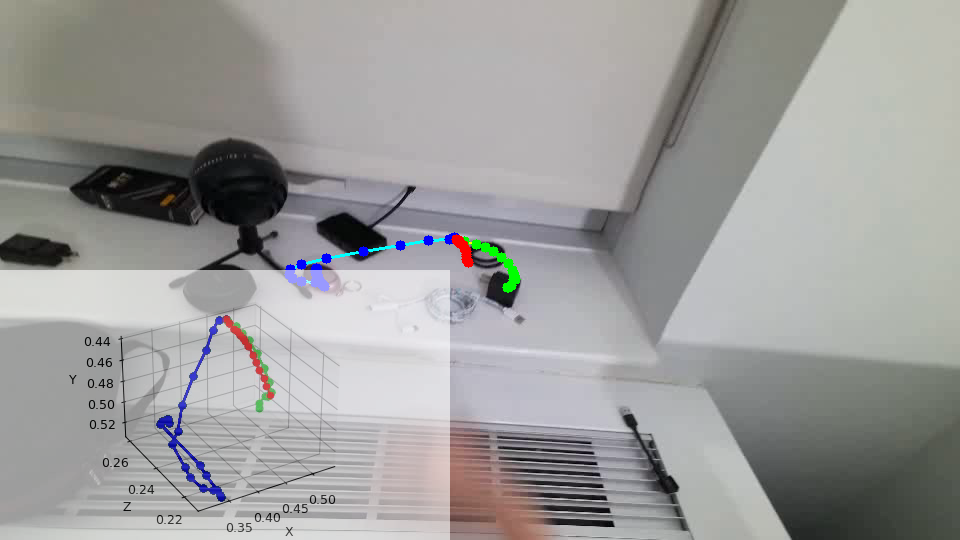} 
\\
\end{tabular}
\captionsetup{font=small,aboveskip=3pt}
\caption{\textbf{Visualization on Unseen data.} For each scene (in a row), we show two examples of the 2D and 3D trajectories on the first frame. The \textcolor{blue}{blue}, \textcolor{green}{green}, and \textcolor{red}{red} trajectory points represent the past observed, future ground truth, and future predictions, respectively. 
}
\label{supfig:vis_unseen}
\vspace{-10pt}
\end{figure*}

In addition to the visualizations of the main paper, we additionally provide some examples as shown in Fig.~\ref{supfig:vis_seen} and~\ref{supfig:vis_unseen}. They show that our method could accurately predict the 3D and 2D hand trajectories in both seen and unseen scenarios. 

\end{appendix}

\end{document}